\documentclass[journal]{IEEEtran}

\usepackage[pagebackref,colorlinks,linkcolor=red,citecolor=green,anchorcolor=blue]{hyperref}
\usepackage[space, nocompress, sort]{cite}
\usepackage{amsmath}
\usepackage{algorithm, algorithmic,graphicx}
\usepackage{float}
\usepackage{bbding}

\ifCLASSINFOpdf
\else
\fi

\usepackage[utf8]{inputenc}
\usepackage{array}
\usepackage{wrapfig}
\usepackage{multirow}
\usepackage{tabularx}
\usepackage{steinmetz}
\usepackage{booktabs}
\usepackage{verbatim}
\usepackage{adjustbox}

\begin{document}
\title{Bridging Component Learning with Degradation Modelling for Blind Image Super-Resolution}

\author{Yixuan~Wu,~Feng~Li,~Huihui~Bai,~Weisi~Lin,\IEEEmembership{~Fellow,~IEEE,}~Runmin~Cong,\IEEEmembership{~Member,~IEEE,} and~Yao~Zhao,~\IEEEmembership{Senior~Member,~IEEE}
\thanks{This work was supported in part by National Key R \& D Program of China (No.2022YFE0200300), the Fundamental Research Funds for the Central Universities (2019JBZ102), the National Natural Science Foundation of China (No. 61972023, 62002014, 62120106009), the Beijing Natural Science Foundation (4222013, L223022), the Beijing Nova Program under Grant (Z201100006820016) , the CAAI-Huawei MindSpore Open Fund and the China Scholarship Council under Grant (202007090046). (Yixuan Wu and Feng Li contributed equally to this work.) Corresponding author: Huihui Bai.}
\thanks{Yixuan Wu, Huihui Bai, Runmin Cong, and Yao Zhao are with the Beijing Key Laboratory of Advanced Information Science and Network Technology, Beijing, 100044, China; and also with the Institute of Information Science, Beijing Jiaotong University, Beijing, 100044, China. Email: \{wuyixuan, hhbai, rmcong, yzhao\}@bjtu.edu.cn}

\thanks{Feng Li is with the School of Computer Science and Information Engineering, Hefei University of Technology, Hefei, 230009, China. Email: fengli@hfut.edu.cn.}

\thanks{Weisi Lin is with the School of Computer Science and Engineering, Nanyang Technological University, 50 Nanyang Avenue, Singapore, 639798. Email: wslin@ntu.edu.sg.}
}

\markboth{Journal of \LaTeX\ Class Files,~Vol.~14, No.~8, August~2015}%
{Shell \MakeLowercase{\textit{et al.}}: Bare Demo of IEEEtran.cls for IEEE Journals}

\maketitle
\begin{abstract}
Convolutional Neural Network (CNN)-based image super-resolution (SR) has exhibited impressive success on known degraded low-resolution (LR) images. However, this type of approach is hard to hold its performance in practical scenarios when the degradation process (\emph{i.e.} blur and downsampling) is unknown. Despite existing blind SR methods proposed to solve this problem using blur kernel estimation, the perceptual quality and reconstruction accuracy are still unsatisfactory. In this paper, we analyze the degradation of a high-resolution (HR) image from image intrinsic components according to a degradation-based formulation model. We propose a components decomposition and co-optimization network (CDCN) for blind SR. Firstly, CDCN decomposes the input LR image into structure and detail components in feature space. Then, the mutual collaboration block (MCB) is presented to exploit the relationship between both two components. In this way, the detail component can provide informative features to enrich the structural context and the structure component can carry structural context for better detail revealing via a mutual complementary manner. After that, we present a degradation-driven learning strategy to jointly supervise the HR image detail and structure restoration process. Finally, a multi-scale fusion module followed by an upsampling layer is designed to fuse the structure and detail features and perform SR reconstruction. Empowered by such degradation-based components decomposition, collaboration, and mutual optimization, we can bridge the correlation between component learning and degradation modelling for blind SR, thereby producing SR results with more accurate textures. Extensive experiments on both synthetic SR datasets and real-world images show that the proposed method achieves the state-of-the-art performance compared to existing methods. The source code is available at \url{https://github.com/Arcananana/CDCN}.
\end{abstract}


\IEEEpeerreviewmaketitle

\section{Introduction}
Single image super-resolution (SR) aims at restoring the high-resolution (HR) image from a low-resolution (LR) image, which is a classical but active task in computer vision. In the past years, various convolutional neural network (CNN) methods have significantly promoted the development of image SR \cite{SRCNN,VDSR,DRRN,LapSRN,EDSR,CARN,RCAN,ESRGAN,SRFBN,HAN} and achieved remarkable progress. Most CNN-based methods assume that an observed LR image is generated by degrading an HR image via a bicubic downsampling kernel. However, when the real degradation differs from such bicubic assumption, these methods can suffer from obvious performance drops. Recent researches~\cite{SRMD,USRNet,UDVD} assume the degradation process of an LR image is known and define the degradation kernel (\emph{e.g.} Gaussian blurs and noises) as prior information to recover the HR image. Though this type of non-blind approaches exhibits better performance than previous bicubic-based methods, in real scenarios, it faces the limitation of versatility when the degradation information is complex or even unknown. To fulfill the need for practical applications, it still inevitably adopts kernel estimation to provide degradation information. 

\begin{figure*}[ht]
\centering
\includegraphics[width=1.0\linewidth]{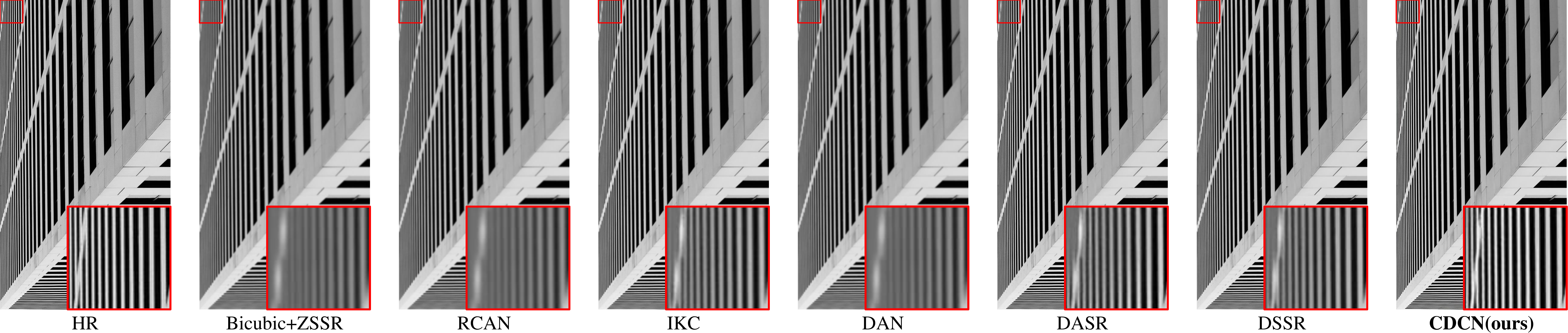}
\caption{SISR results on \emph{''img\_011"} in Urban100 \cite{urban100} with scale factor 2. The HR image is first blurred by an isotropic Gaussian kernel with kernel width set to 1.8 and then downsampled by a bicubic kernel. Equipped with the component decomposition and degradation-driven learning strategy, the proposed CDCN succeed to recover sharp and realistic details,  outperforming other SOTA blind SR methods including IKC \cite{IKC}, DAN \cite{DAN}, DASR \cite{DASR} and DSSR \cite{DSSR}. }
\label{fig:1}
\end{figure*}

To overcome this issue, blind SR methods~\cite{kernelgan,IKC,kmsr,DAN,MANet,FKP,DSSR,DASR,KOALAnet,realsr,moesr} propose to recover the HR image from an unknown degraded LR image. Some of them~\cite{kernelgan,kmsr,FKP,realsr,moesr,MANet} divide this problem into degradation estimation and kernel-based SR reconstruction. Bell-Kligler \emph{et al.}\cite{kernelgan} introduce an image-specific internal-GAN to estimate the SR kernel according to the distribution of patches across scales of the LR image. Zhou and S{\"u}sstrunk\cite{kmsr} build a large blur kernel pool from real photographs and then employ them to create a paired LR-HR dataset for training. In~\cite{MANet}, a mutual affine network (MANet) is presented to estimate spatially variant kernels from the LR observation for every HR image pixel. These methods highly rely on explicit kernel estimation. Actually, obtaining accurate kernels from arbitrary images is difficult, where possible estimation errors can cause visually unpleasant artifacts in SR results. Some methods~\cite{IKC,DAN} perform iterative kernel estimation and conditional SR reconstruction in an end-to-end framework, which take advantage of intermediate SR results to alleviate the kernel mismatch problem. Yet, the time consumption caused by multiple iterations cannot be ignored by these two methods. Wang~\emph{et al.}~\cite{DASR} propose an unsupervised learning scheme to learn degradation representations for blind SR, which realizes fast reconstruction speed but weak accuracy. 

The above methods directly model the degradation process from LR images while ignoring ambiguity from downsampling operation (e.g. bicubic downsampling), which increases the difficulty of accurate kernel estimation and further reduces the possibility of vivid reconstruction. To avoid the adverse effects of degradation estimation, Li~\emph{et al.}~\cite{DSSR} propose to alternatively optimize the HR image detail and latent structural context using a recurrent framework for blind SR (DSSR), which produces acceptable results without degradation prior incorporation. Nevertheless, there are main two drawbacks to this method: 1) DSSR simply supervises the residue between the predicted HR image and bicubic-upsampled image to optimize the HR detail, where the structure component is implicitly optimized in latent space by the SR supervision. This manner ignores the influence of degradation for SR reconstruction, which is sub-optimal to generate satisfactory results. 2) DSSR also adopts an iterative learning scheme to pursue robust performance. But, its corresponding training process or testing process is converged in several iterations, which seems to reach saturation of model performance. In fact, whether the local details or perceptual quality of the SR results can be further improved (see Fig.~\ref{fig:1}).

All these mentioned observations motivated us to consider two issues: 1) How to avoid the adverse effects of kernel estimation while modelling degradation information. 2) Is there any more effective image detail and structure optimization scheme for robust blind SR performance. To this end, in this paper, we first introduce a degradation-based formulation model that denotes the degradation kernel of an HR image as a combination of the blur kernel, bicubic kernel, and standard downsampler. According to this formulation model, we analyze the degradation of an HR image from its intrinsic components (see~Sec.~\ref{subsection:3-A}). Therefore, we propose a components decomposition and co-optimization network (CDCN), which bridges component learning and degradation modelling for blind SR. CDCN starts from simply LR image decomposition in feature space to obtain initial LR detail and structure components. Then, considering that the two components represent different image characteristics but are interdependent, we present the mutual collaboration block (MCB) to exploit the relationship between both components and learn more comprehensive representations conditioned on each other, achieving a collaborative optimization process. Based on our degradation-based formulation, we present a degradation-driven learning strategy to jointly recover the HR image detail and structure. Moreover, we leverage a multi-scale fusion module that combines the learned detail and structure features for final HR image restoration. Experimental results demonstrate that CDCN outperforms recent state-of-the-art blind SR methods on both synthetic datasets and real-world images.

We summarize the contributions of this work as follows:

\begin{itemize}
	\item We analyze the degradation of an HR image from image intrinsic components and propose a components decomposition and co-optimization network (CDCN), which bridges component learning and degradation modelling for blind SR.
	
	\item We propose the mutual collaboration block (MCB), which exploits the relationship between both image structure and detail components, enabling collaborative optimization conditioned on each other. 
	
	\item We propose a degradation-driven learning strategy to jointly perform HR detail and structure recovery. Such a strategy maintains consistency with the degradation process and provides a new perspective for image component learning under blind settings.
\end{itemize}

\section{Related Work}%
\subsection{Single Degradation Image Super-Resolution}
CNN-based image SR with single degradation aims at super-resolving LR images that are degraded from HR images with predefined single degradation, \emph{i.e.} bicubic downsampling. This type of approaches constructs paired training data based on a bicubic kernel and investigates various CNN architectures to learn the LR-to-HR mapping. Early methods~\cite{SRCNN,VDSR,DRCN,DRRN,Memnet} conduct pre-interpolation on LR images using bicubic upsampling and learn to refine the upscaled HR images. Due to much calculation on large-resolution features in CNN layers, these methods face the challenges of computational complexity and time cost. To address this problem, some methods~\cite{LapSRN,SRdensenet,CARN,MRFN,SRGAN,EDSR,ESPCN,RDN} propose to directly learn the feature representations from the LR input and amplify learned features onto HR space at the end of networks using transposed convolution~\cite{LapSRN,SRdensenet,MRFN,filternet} or sub-pixel layer \cite{CARN,ESPCN,SRGAN}. Ledig~\emph{et al.} \cite{SRGAN} propose to train a generative adversarial network (GAN) with perceptual loss and adversarial loss for photo-realistic image SR reconstruction. Wang~\emph{et al.}~\cite{ESRGAN} propose enhanced SR generative adversarial networks (ESRGAN) by introducing local residual dense connection, relativistic discriminator, and deeper architecture in \cite{SRGAN}, which significantly boosts the SR performance. In~\cite{DBPN}, Haris~\emph{et al.} exploit iterative feature upsampling and downsampling operations to provide an error feedback mechanism, constituting a deep back-projection network (DBPN) for image SR. In \cite{prosr}, a multipath recursive residual learning framework is introduced for several classical image restoration tasks including SISR. Liu~\emph{et al.}~\cite{isrn} perform iterative SR reconstruction and downsampling to gradually decrease the distance on the HR space by adjusting the distance on the LR space. Recently, the attention mechanism~\cite{RCAN,SAN,HAN,fine,DRSAN} is widely used in various image SR networks to exploit more informative features for performance improvement.

\subsection{Non-Blind Super-Resolution with Multiple Degradations}
Except above single degradation SR methods for non-bind bicubic downsampling setting, many other non-blind methods~\cite{SRMD,DPSR,USRNet,UDVD,MZSR} aim at super-resolving the LR image under multiple degradations using a single model. These methods assume a fixed known but more complex degradation process and unify the degradation prior with LR input to train a single SR model with more practicalities. In~\cite{SRMD}, Zhang~\emph{et al.} propose a dimensionality stretching strategy to produce the degradation map that is concatenated with an LR image for SR reconstruction, which can solve blur, noise, and bicubic downsampling degradation in a single model. The authors later present a deep plug-and-play SR (DPSR) framework~\cite{DPSR}, which combines deblurring methods and a half quadratic splitting optimization algorithm to solve arbitrary uniform blur kernels. In \cite{USRNet}, an unfolding SR network (USRNet) is proposed to alternatively optimize a data sub-problem and a prior sub-problem for multiple degradations. Xu \emph{et al.} \cite{UDVD} adopt the pre-defined degradation map as an additional input and employ dynamic convolution to enhance the effectiveness of networks for variational degradations. Hussein \emph{et al.}~\cite{closed-form} introduce a closed-form correction filter to convert an LR image to match the bicubic degraded one, thus the existing model for bicubic degradation can super-resolve the converted image. Shocher \emph{et al.} \cite{ZSSR} present a zero-shot SR (ZSSR) method which exploits the internal recurrence of information inside a single image across different scales to perform image-specific SR. Soh \emph{et al.} \cite{MZSR} first exploit external data to learn a generic initial parameter of the SR model and then employ meta-learning to make the model adapt to given images, which significantly improves the inference speed for ZSSR.  

\subsection{Blind Super-Resolution}
For blind SR, the degradation from an HR image to an LR one is unknown. Under such blind setting, existing methods mainly pursue estimating accurate degradation kernels from LR observations and conditionally reconstructing HR images based on estimated kernels. Cornillere \emph{et al.} \cite{blindSR} propose a degradation-aware SR network that generates both the HR image and the blur kernel. A kernel discriminator is also designed in \cite{blindSR} to refine their results. Gu \emph{et al.} \cite{IKC} introduce an iterative kernel correction (IKC) method to iteratively correct inaccurate blur kernels by observing the intermediate SR results. Bell-Kligler \emph{et al.} \cite{kernelgan} use an internal generative adversarial network (GAN) to estimate degradation kernel from LR images at test time. Luo \emph{et al.} \cite{DAN} adopt an alternating optimization algorithm, which can estimate blur kernels and restore SR images iteratively. Wang \emph{et al.} \cite{DASR} extract abstract representations to distinguish different degradation with contrastive learning and super-resolve LR images using the discriminative representations. Liang~\emph{et al.}\cite{FKP} propose a flow-based kernel prior (FKP) method, which models the kernel distribution by learning an invertible mapping between the kernel space and a tractable latent space. In \cite{MANet}, a mutual affine network (MANet) with a moderate receptive field is proposed to maintain the locality of degradation for spatially variant kernel estimation. Kim \emph{et al.} \cite{KOALAnet} use a downsampling network to estimate spatially variant kernels. The predicted kernels are then leveraged as local filtering operations for SR features modulation. Jo \emph{et al.} \cite{adatarget} propose to use an adaptive target rather than the GT target, which can alleviate the ill-posedness of blind SR problem and encourage sharp SR reconstruction. He \emph{et al.} \cite{SRDRL} introduce a degradation reconstruction loss to guide the SR network to capture the degradation-wise differences between HR images and SR results. Li \emph{et al.} \cite{DSSR} introduce a recurrent alternative optimization SR network to tackle the blind SR problem from structure and detail perspectives without explicit degradation incorporation. MoESR \cite{moesr} utilizes a mixture of experts to board the degradation space and trains an expert for a specific kernel.

\subsection{Component Learning-based Super-Resolution}\label{section:2-D}
In image SR, the key idea of component learning is to optimize the intrinsic image components of LR images to facilitate SR reconstruction. Shi~\emph{et al.}~\cite{SPSR} impose the base image content, global boundary context, and residual context as complementary information and jointly learn the three contexts with a multitask framework to address the SR problem. Yang~\emph{et al.}~\cite{DEGREE} take the LR image and its edge map as input and jointly infer their HR versions by a recurrent residual SR network. Xie~\emph{et al.}~\cite{CLSR} introduce a convolutional sparse coding (CSC)-based decomposition method to separate the LR input into residual and smooth components. The residual component is represented by a shallow CNN and then re-combined with the interpolated smooth component to predict the HR image. Nazeri~\emph{et al.}~\cite{edgeinform} use a two-stage network to sequentially recover HR edges and textures. Ma~\emph{et al.}~\cite{SPSR2} propose a structure-preserving SR network (SPSR) which exploits gradient maps as auxiliary context to guide the recovery of HR images. In~\cite{TDPN}, Cai~\emph{et al.} propose to decompose the textures and details from ground-truth HR images to guide the network to learn textures and details in observed LR images for better perceptual quality.

\section{Proposed Method}\label{section:3}

\subsection{Degradation-based Problem Formulation}\label{subsection:3-A}

In the typical SISR framework, the degradation model between an HR image $I_{HR}$ and its corresponding LR image $I_{LR}$ can be formulated as follows:
\begin{equation}
I_{LR}=\left(I_{HR}\otimes{k_d}\right)\downarrow_{s} \label{eq:1}
\end{equation}
where $k_d$ denotes the degradation kernel, $\otimes$ denotes the convolution operation, $\downarrow_{s}$ represents a downsampling operation with a scale factor $s$. In the blind SR problem, most methods \cite{IKC,DAN,DASR} adopt the combination of isotropic Gaussian kernels or anisotropic Gaussian kernels together with additive white Gaussian noise as the degradation kernel $k$ and bicubic downsampler as the downsampling operation, respectively. Existing methods \cite{IKC,DAN} usually only take the Gaussian blur kernel ${k_g}$ into consideration and treat the blind SR problem as an optimization problem:
\begin{equation}
\begin{cases}
{k_g}=\mathcal{E}\left({I_{LR}}\right) \\
\theta_{sr}=\mathop{\arg\min}\limits_{\theta_{sr}}\lVert{I_{HR}}-\mathcal{SR}\left(I_{LR};{k_g};\theta_{sr}\right)\rVert_{1} \label{eq:2}\\
\end{cases}
\end{equation}
where $\mathcal{E}\left(\cdot\right)$ is the function that estimates the Gaussian blur kernel ${k_g}$ from $I_{LR}$, $\mathcal{SR}\left(\cdot\right)$ is the function that restores $I_{HR}$ from $I_{LR}$, $\theta_{sr}$ is the parameter of $\mathcal{SR}\left(\cdot\right)$. However, estimating a single blur kernel from the input LR image for the whole image restoration may cause two intrinsic problems: 1) The input LR image is blurred by the combination of the Gaussian blur kernel ${k_g}$ and the downsampling operation, which increases the difficulty of accurate Gaussian blur kernel estimation. 2) The degradation process is ill-posed, which means a degraded patch may correspond to several blur kernels. In fact, when applying blur kernels to an HR image, not only blur kernels themselves but also the HR image plays a decisive role in generating the LR image.

To tackle these problems, since the bicubic degradation can be approximated by setting a proper blur kernel in Eq.(\ref{eq:1}) \cite{kernelgan,USRNet}, in this paper, we view the bicubic downsampler as a combination of bicubic kernel and standard downsampler. For the noise-free with Gaussian kernels degradation model, we re-formulate Eq.(\ref{eq:1}) as

\begin{equation}
I_{LR}=I_{HR}\otimes{k_g}\otimes{k_b}\downarrow_{s_d} \label{eq:3}
\end{equation}
where the input HR image $I_{HR}$ is first blurred by the combination of the Gaussian blur kernel ${k_g}$ and the bicubic kernel ${k_b}$. Then it is downsampled by the standard s-fold downsampler $s_d$, which only keeps the upper-left pixel for each distinct $s \times s$ patch.  

Subsequently, we propose to model the degradation process from image intrinsic components. According to Eq.~(\ref{eq:3}), we define an HR image $I_{HR}$ degraded by a Gaussian kernel $k_g$ as the structure component $I_s$, and the lost information as the detail component $I_d$. In this condition, we can obtain an LR image $I_{LR}$ by downsampling $I_s$ with
bicubic kernel $k_b$. The decomposition process can be expressed as
\begin{equation}
\begin{cases}
I_{s}=I_{HR}\otimes{k_g} \\
I_{d}=I_{HR}-I_{s} \label{eq:4}\\
I_{LR}=I_{s}\otimes{k_b}\downarrow_{s}
\end{cases}
\end{equation}

Consequently, the optimization problem in Eq.~(\ref{eq:2}) can be decomposed into two separated optimization problems:
\begin{equation}
\begin{cases}
\theta_s=\mathop{\arg\min}\limits_{\theta_s}\lVert I_{s} -\mathcal{S}\left(I_{LR};\theta_{s}\right)\rVert_{1}\label{eq:5} \\
\theta_d=\mathop{\arg\min}\limits_{\theta_d}\lVert I_{d} -\mathcal{D}\left(I_{LR};\theta_{d}\right)\rVert_{1}  \\
\end{cases}
\end{equation}
where $\mathcal{S}\left(\cdot\right)$ is the function that restores the structures of $I_{HR}$ from $I_{LR}$. $\theta_s$ is the learned parameter by $\mathcal{S}\left(\cdot\right)$. It is worth noting that since the $I_{LR}$ is directly bicubic downsampled from $I_{S}$, the $\mathcal{S}\left(\cdot\right)$ is equivalent to $\mathcal{SR}\left(\cdot\right)$ in the traditional SISR tasks. $\mathcal{D}\left(\cdot\right)$ is the function that estimates the details of $I_{HR}$ from $I_{LR}$. $\theta_d$ is the learned parameter by $\mathcal{D}\left(\cdot\right)$.

\begin{figure*}[t]
\centering
\includegraphics[width=1.0\linewidth]{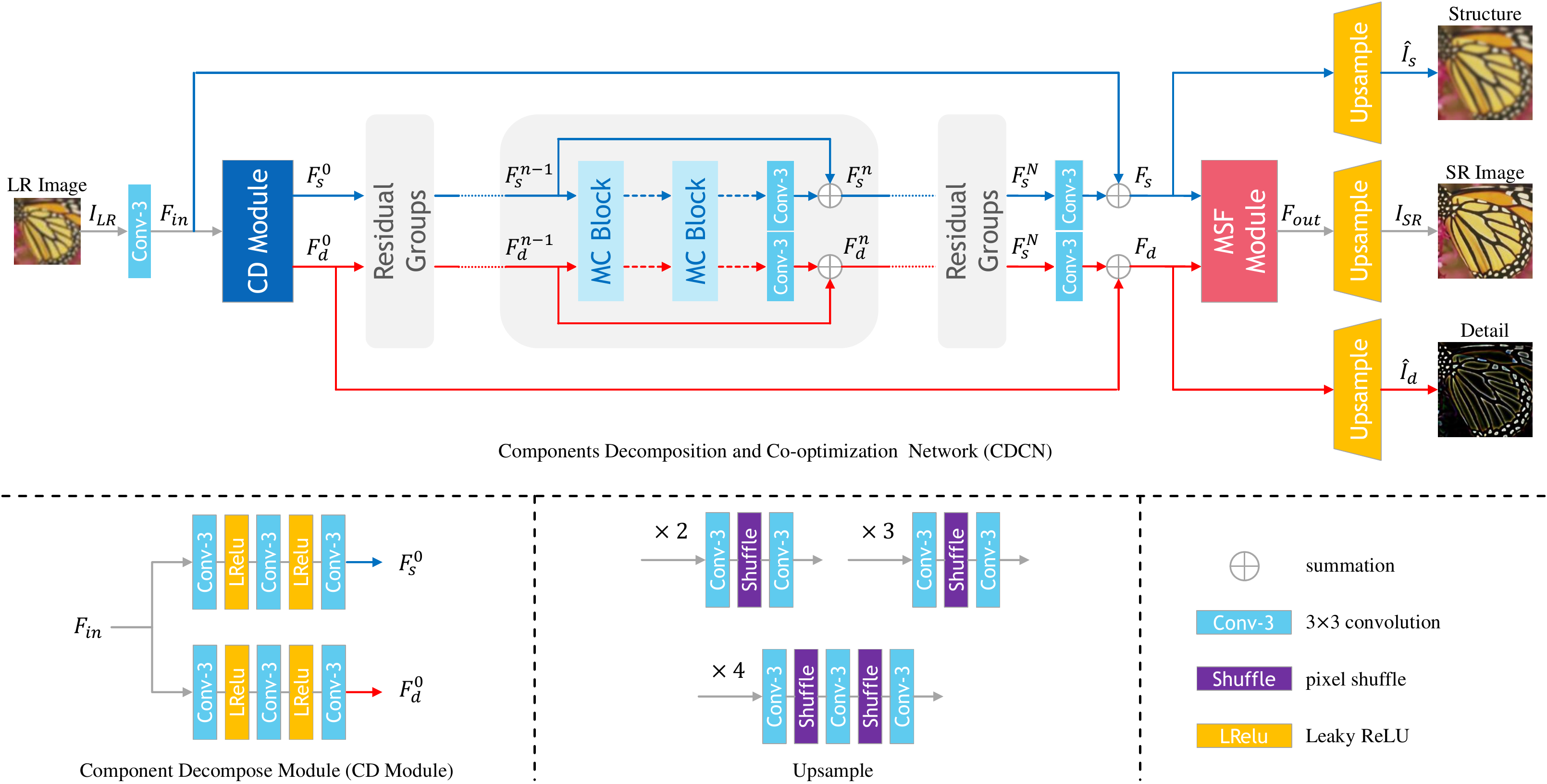}
\caption{The architecture of the proposed components decomposition and co-optimization network (CDCN). \textcolor{blue}{Blue} arrows represent feature connections of the structure component. \textcolor{red}{Red} arrows represent feature connections of the detail component.}
\label{fig:2}
\end{figure*}

Considering that the detail component can reveal informative regions to help structure inferring and the structure component can help to understand image context, thus it is sub-optimal to separately optimize the detail or structure components for accurate SR reconstruction. To better exploit the collaboration between structure and detail components, we propose to continuously enhance the structure and detail feature representations in a mutual complementary way, resulting in a collaborative optimization processing for blind SR. To this end, the objective of this work can be further rewritten as follows:

\begin{equation}
\begin{cases}
\hat{I}_{s}=\mathcal{S}\left(I_{LR};\theta_{s};\theta_{d}\right)\\
\hat{I}_{d}=\mathcal{D}\left(I_{LR};\theta_{s};\theta_{d}\right)\\
\theta_s=\mathop{\arg\min}\limits_{\theta_s}\lVert I_{s} -\hat{I}_{s}\rVert_{1} \label{eq:6}\\
\theta_d=\mathop{\arg\min}\limits_{\theta_d}\lVert I_{d} -\hat{I}_{d}\rVert_{1}  \\
\end{cases}
\end{equation}
where $\hat{I}_{s}$ and $\hat{I}_{d}$ are the predicted structure and detail components using $I_{LR}$. In contrast to Eq.~(\ref{eq:5}), the $\mathcal{S}\left(\cdot\right)$ and $\mathcal{D}\left(\cdot\right)$ not only concern the information from $I_{LR}$ but also concern the information from each other. Thus the difficulty of detail estimation can be eased by the structures and the quality of structure restoration can be enhanced by the detail information. 

With optimized structure and detail components, we finally combine them to synthesize the SR image $I_{SR}$, which is completed by the fusion function $\mathcal{F}\left(\cdot\right)$. The SR image $I_{SR}$ can be generated by the following process:
\begin{equation}
	\begin{aligned}
		I_{SR}&=\mathcal{F}\left(\hat{I}_{s};\hat{I}_{d};\theta_{f}\right)\\
		&=\mathcal{F}\left(\mathcal{S}\left(I_{LR};\theta_{s};\theta_{d}\right);\mathcal{D}\left(I_{LR};\theta_{s};\theta_{d}\right);\theta_{f}\right)\\
		&=\mathcal{F}_{SR}\left(I_{LR};\theta_{s};\theta_{d};\theta_{f}\right)\\
	\end{aligned}
\end{equation}
where $\mathcal{F}_{SR}\left(\cdot\right)$ is the whole SR function of the proposed scheme and $\theta_f$ is the learned parameter by $\mathcal{F}\left(\cdot\right)$. We minimize the distance between $I_{HR}$ and generated $I_{SR}$ to facilitate our SR reconstruction. Such a process can be formulated as follows:
\begin{equation}
	\begin{aligned}
		\theta_f&=\mathop{\arg\min}\limits_{{\theta_f}}\lVert I_{HR}-I_{SR}\rVert_{1} \\
		&=\mathop{\arg\min}\limits_{{\theta_f}}\lVert I_{HR}-\mathcal{F}_{SR}\left(I_{LR};\theta_{s};\theta_{d};\theta_{f}\right)\rVert_{1} \\
	\end{aligned}
\end{equation}

\subsection{Network Structure}
\textbf{Overall framework.} The proposed components decomposition and co-optimization network (CDCN) aims at the tackling blind SR problem by modelling the degradation process from a component decomposition perspective, where its architecture is shown in Fig.~\ref{fig:2}. The CDCN is constructed by three parts: a component decomposition module (CDM), cascaded residual groups (RG) consisting of multiple mutual collaboration blocks (MCB), and a multi-scale fusion module (MSFM). 

Given an LR image $I_{LR}$ of size $C\times H \times W$, where $C$ denotes the number of channels, $H$ and $W$ denote the height and width of $I_{LR}$. We first feed $I_{LR}$ into a $3 \times 3$ convolutional layer to extract the shallow feature $F_{in}$, which serves as the input of our CDM for component decomposition. In CDM, we obtain the initial detail and structure components by using two individual feature extraction blocks to directly learn the corresponding features from $F_{in}$. This process can be formulated as follows:
\begin{equation}
F^0_s, F^0_d=\mathcal{C}(F_{in})
\end{equation}
where $\mathcal{C}\left(\cdot\right)$ represents the component decomposition function to produce the structure component $F^0_s$ and detail component $F^0_d$.

After that, to exploit the relationship between both components, we design the mutual collaboration block (MCB), which performs detail and structure information interaction and learns more comprehensive representations conditioned on each other. By forming multiple MCBs in a stacked manner, we can construct the residual groups (RG) to learn powerful feature representations. Supposing there are $N$ RGs in CDCN and each RG contains $M$ MCBs, for the $m$-th MCB in the $n$-th RG, the process of structure and detail component learning can be expressed as  
\begin{equation}
\begin{aligned}
F_s^{m,n}, F_d^{m,n}&=\mathcal{B}_{m,n}(F_s^{m-1,n},F_d^{m-1,n}) \\
&=\mathcal{B}_{m,n}(...(\mathcal{B}_{1,n}(F_s^{n-1},F_d^{n-1}))...)
\end{aligned}
\end{equation}
where $\mathcal{B}(\cdot)$ represents the function of MCB, $(F_s^{m-1,n}, F_d^{m-1,n})$ and $(F_s^{m,n},F_d^{m,n})$ are the input and output of the $m$-th MCB in the $n$-th RG, respectively. With local residual connections, the output of the $n$-th RG can be obtained as follows 
\begin{equation}
\begin{cases}
F_s^n=F_s^{n-1}+{W_s^{n-1}}F_s^{M,n-1} \\
F_d^n=F_d^{n-1}+{W_d^{n-1}}F_d^{M,n-1} \\
\end{cases}
\end{equation}
where $(F_s^n,F_d^n)$ and $(F_s^{n-1},F_d^{n-1})$ are the input and output of the $n$-th RG respectively. $W_s^{n-1}$ and $W_d^{n-1}$ are the weight sets of the two convolutional layers at the tail of $(n-1)$-th RG. Such a process can also be abbreviated as
\begin{equation}
\begin{aligned}
F_s^n, F_d^n&=\mathcal{G}_{n}(F_s^{n-1},F_d^{n-1})\\
&=\mathcal{G}_n(\mathcal{G}_{n-1}(...(\mathcal{G}_1(F_s^0,F_d^0))...))
\end{aligned}
\end{equation}
where $\mathcal{G}\left(\cdot\right)$ represents the function of RG. 

\begin{figure*}[t]
\centering
\includegraphics[width=1.0\linewidth]{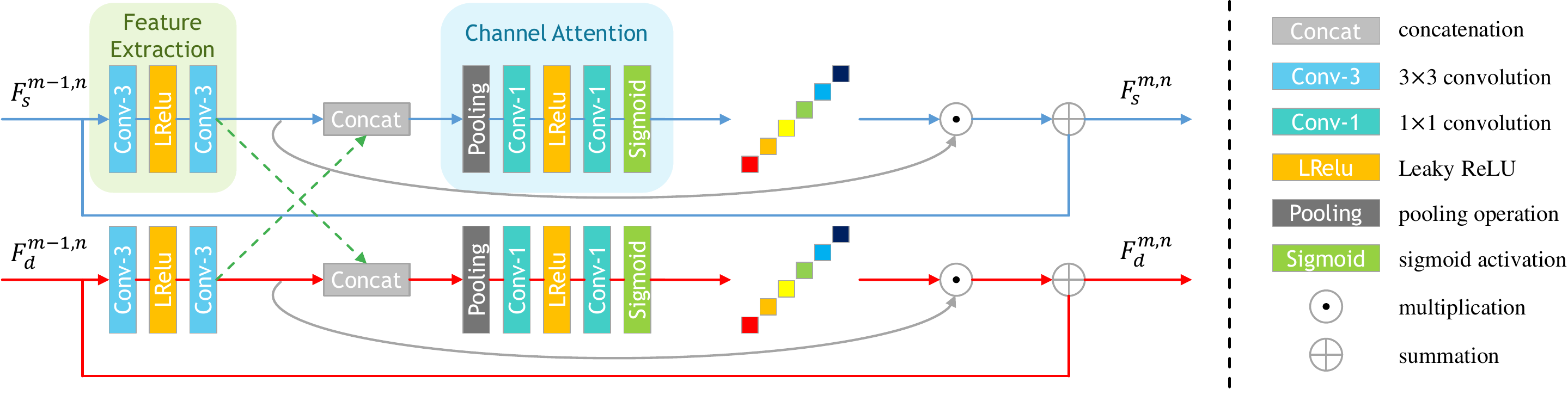}
\caption{The proposed mutual collaboration block (MCB). \textcolor{blue}{Blue} arrows represent feature connections of structure component. \textcolor{red}{Red} arrows represent feature connections of detail component.}
\label{fig:3}
\end{figure*}

The very shallow LR feature $F_{in}$ contains abundant structural contexts, which are helpful for better structure modelling. Therefore, after stacking $N$ RGs, we design an asymmetrical global residual connection by
\begin{equation}
\begin{cases}
F_s=F_{in}+W_{s}F_s^{N}\\
F_d=F_d^{0}+W_{d}F_d^{N}\\
\end{cases}
\end{equation}
where $W_s$ and $W_d$ are the weight sets learned by the two convolutional layers after $N$ RGs.
Then we use MSFM to fuse the features of structure and detail components, which outputs the final feature $F_{out}$
\begin{equation}
F_{out}=\mathcal{F}(F_s, F_d)
\end{equation}
where $\mathcal{F}\left(\cdot\right)$ is the function of MSFM.

Finally, we use three individual upsample modules to project the LR features $F_s$, $F_d$, and $F_{out}$ onto HR space and generate corresponding predicted HR structure image $\hat{I}_s$, predicted HR detail image $\hat{I}_d$, and final SR result $I_{SR}$. The architecture of upsample modules is illustrated in Fig.~\ref{fig:2} (bottom middle).

\subsection{Mutual Collaboration Block}
The detail and structure components represent different image characteristics but are interdependent. On the one hand, the structure component can help understand contextual information and find the location of the detail component. On the other hand, the detail component can reveal the informative region in the image, thus assisting the reconstruction of the structure component. Consequently, we propose the mutual collaboration block (MCB) to better learn the structure and detail features of images. As shown in Fig.~\ref{fig:3}, For the $m$-th MCB in the $n$-th RG, there are structure component $F_s^{m-1,n}$ and detail component $F_d^{m-1,n}$ as the input. We first use two individual stacked convolutional layers to perform feature extraction on $F_s^{m-1,n}$ and $F_d^{m-1,n}$.
\begin{equation}
\begin{cases}
X_s^{m,n}=W_{s2}^{m,n}\delta(W_{s1}^{m,n}F_s^{m-1,n}) \\
X_d^{m,n}=W_{d2}^{m,n}\delta(W_{d1}^{m,n}F_d^{m-1,n}) \\
\end{cases}
\end{equation}
where $X_s^{m,n}$ and $X_d^{m,n}$ are the extracted features. $[W_{s1}^{m,n}$, $W_{s2}^{m,n}]$ and $[W_{d1}^{m,n},W_{d2}^{m,n}]$ are the corresponding learned weight sets for $X_s^{m,n}$ and $X_d^{m,n}$ in MCB, respectively. $\delta(\cdot)$ represents LeakyReLU\cite{leakyrelu} activation function.

Next, to exploit the relationship between structure and detail features, we concatenate $X_s^{m,n}$ and $X_d^{m,n}$ to make a fused feature representation $X^{m,n}$ and conduct parallel channel-wise attention to capture the inter-channel dependencies. Thus, we can produce two attention maps $A_s^{m,n}$ and $A_d^{m,n}$
\begin{equation}
    \begin{cases}
    A_s^{m,n}={\mathcal{CA}_s}(X^{m,n}) \\
    A_d^{m,n}={\mathcal{CA}_d}(X^{m,n}) \\ 
    \end{cases}
\end{equation}
where $\mathcal{CA}_s(\cdot)$ and $\mathcal{CA}_d(\cdot)$ denote the functions of channel attention. We further use $A_s^{m,n}$ and $A_d^{m,n}$ to re-weight the structure and detail features respectively, which enforces the network focus on more informative features. By integrating such a mechanism in a residual block, the output of MCB can be formulated as
\begin{equation}
    \begin{cases}
    F_s^{m,n}=F_s^{m-1,n}+A_s^{m,n}X_s^{m,n} \\ 
    F_d^{m,n}=F_d^{m-1,n}+A_d^{m,n}X_d^{m,n} \\
    \end{cases}
\end{equation}
where $F_s^{m,n}$ and $F_d^{m,n}$ denote the resulting structure and detail features. Equipped with MCB, we can effectively exploit the interaction between image structure and detail components, learning more comprehensive representations based on each other, and achieving a collaborative optimization process. The effectiveness of the proposed MCB is discussed in Section \ref{subsection:4-B}.

\begin{figure*}[t]
\centering
\includegraphics[width=1.0\linewidth]{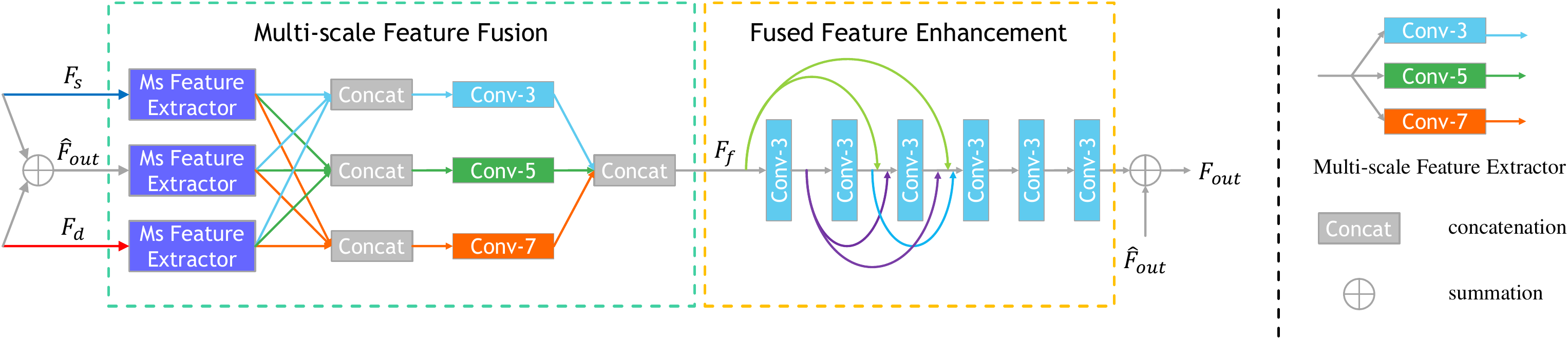}
\caption{The proposed multi-scale fusion module (MSFM) in CDCN, where ``conv-3'', ``conv-5'', and ``conv-7'' denote $3\times3$, $5\times5$, and $7\times7$ convolutions, respectively.}
\label{fig:4}
\end{figure*}

\subsection{Multi-scale Fusion Module}
The goal of our multi-scale fusion module (MSFM) is to ensure that the two component features $F_s$ and $F_d$ can be well combined, thus improving the final reconstruction quality. In MSFM, we first take $F_s$, $F_d$, and their summation $\hat{F}_{out}$ as the input for multi-scale feature extraction. As shown in Fig.~\ref{fig:4}, we leverage three parallel convolutions with kernel sizes of $3\times 3$, $5\times 5$, and $7\times 7$, as the multi-scale feature extractor. Therefore, for each input, we can produce three types of features. Taking $F_s$ as example, after such multi-scale feature extraction, we can generate three outputs: $F^{3\times 3}_s$, $F^{5\times 5}_s$, and $F^{7\times 7}_s$.

As for feature fusion, we conduct cross-branch concatenation to aggregate multi-scale information. We use another three parallel convolutional filters with feature concatenation to output the fused feature. The whole multi-scale feature fusion of MSFM can be simply expressed as 
\begin{equation}
    F_{f}=Fusion(F_s, F_d)
\end{equation}    
where $Fusion(\cdot)$ denotes the fusion function. Moreover, in MSFM, we append six densely connected $3\times 3$ convolutional layers with residual connection to enhance the fused feature $F_{f}$
\begin{equation}
    F_{out}=\hat{F}_{out}+E(F_f) \\
\end{equation}
where $E(\cdot)$ denotes the function for feature enhancement. $F_{out}$ is the final output of MSFM.

\subsection{Degradation-driven Learning Strategy}\label{subsection:3-C}
Considering the importance of the degradation process in the blind SR problem, we present a degradation-driven learning strategy to exploit the spatial relation of the degradation kernel and SR reconstruction. Specifically, according to Eq.~(\ref{eq:4}), the kernel-degraded HR image $I_{s}$ is defined as the structure label. We regard the input LR image $I_{LR}$ as a bicubic-blurred and $s$-fold downsampled one on $I_{s}$, \emph{i.e.} $I_{LR}=I_{s}\otimes{k_b}\downarrow_{s}$. Therefore, the estimated structure image $\hat{I}_{s}$ can be represented by $\mathcal{S}\left(I_{s}\otimes{k_b}\downarrow_{s};\theta_{s};\theta_{d}\right)$ (equal to Eq.~(\ref{eq:6})). Thus, the loss function for structure component optimization can be formulated as 
\begin{equation}
\begin{aligned}
   L_{structure}&=\lVert I_{s} -\hat{I}_{s}\rVert_{1}\\
   &=\lVert I_{s} -\mathcal{S}\left(I_{s}\otimes{k_b}\downarrow_{s};\theta_{s};\theta_{d}\right)\rVert_{1}\\
   \label{eq:22}
\end{aligned}
\end{equation}

Besides, the residue between structure label $I_s$ and ground truth $I_{HR}$ is considered as the detail label whose supervision concentrates on lost information caused by blur kernel $k_g$:
\begin{equation}
\begin{aligned}
   L_{detail}&=\lVert I_{d} -\hat{I}_{d}\rVert_{1}\\
   &=\lVert I_{HR}-I_{HR}\otimes{k_g} -\hat{I}_{d}\rVert_{1}\\
\end{aligned}
\end{equation}

The supervision of both detail and structure components drives the network to catch the degradation effect of both blur kernel $k_g$ and bicubic kernel $k_b$, leading to vivid image restoration in the blind SR problem.

In addition, we introduce an SR reconstruction loss to make sure the SR reconstruction result $I_{SR}$ under the supervision of the label $I_{HR}$:
\begin{equation}
L_{SR}=\lVert I_{HR} -I_{SR}\rVert_{1}\\
\end{equation}
The overall loss function to train CDCN can be formulated as follows:
\begin{equation}\label{loss}
    L={L_{structure}}+{L_{detail}}+{L_{SR}}\\
\end{equation}

The degradation-driven learning strategy enables our network to perform 	HR detail and structure recovery and SR reconstruction while maintaining consistency with the degradation process for blind SR. 

\section{Experiments}
\subsection{Datasets and Implementation Details}\label{subsection:4-A}
\textbf{Datasets.} We synthesize the training HR-LR image pairs according to Eq.(\ref{eq:3}) with a bicubic downsampler. For the isotropic Gaussian blur kernels, following \cite{IKC,DAN,DASR,DSSR}, the kernel size is set at $21 \times 21$, and the kernel width ranges are set to $[0.2,2.0]$, $[0.2, 3.0]$ and $[0.2, 4.0]$ for SR scale factors 2, 3 and 4 respectively. For more general degradation with anisotropic Gaussian blur kernels, following \cite{kernelgan}, the kernel size is set at $11 \times 11$ with random lengths $\lambda_{1},\lambda_{2} \sim U(0.6,5)$ independently distributed for each axis, rotated by a random angle $\theta \sim U(-\pi,\pi)$. Then we further apply uniform multiplicative noise (up to 25$\%$ of each pixel value of the kernel) and normalize it to sum to one to make it deviate from a regular Gaussian kernel.

We use 800 training images in DIV2K \cite{div2k} and 2650 training images in Flickr2K \cite{flickr2k}, which is a total of 3450 HR images as our training set. For testing, we evaluate SR performance of degradation with isotropic Gaussian blur kernels on 5 benchmark datasets: Set5 \cite{set5}, Set14 \cite{set14}, BSD100 \cite{b100}, Urban100 \cite{urban100} and Manga109 \cite{manga109}. Specifically, we use \emph{Gaussian8} kernel setting in \cite{IKC}, which uniformly selects 8 kernels whose width range are from $[0.80, 1.60]$, $[1.35, 2.40]$ and $[1.80, 3.20]$ for scale factor 2, 3 and 4 respectively. As for degradation with anisotropic Gaussian blur kernels, we use benchmark dataset DIV2KRK in \cite{kernelgan}. For performance evaluation, we use peak signal-to-noise ratio (PSNR) and structural similarity (SSIM) on the Y channel of transformed YCbCr space as metrics.

\textbf{Parameters.} In the proposed network, the number of the RGs and the number of MCBs per RG are 5 and 10 respectively. The number of filters in the intermediate layer is set to 64. We use LeakyRelu \cite{leakyrelu} as the activation function. 

\textbf{Training settings.} During the training, the patch size is $64 \times 64$ for all scale factors. The batch size is 16. Training dataset augmentation is performed by random horizontal flips and 90-degree rotations. We use Adam \cite{adam} as optimizer with $\beta_1 = 0.9, \beta_2 = 0.99 $ and initial learning rate $ 2 \times 10^{-4}$. All the models are trained for $5 \times 10^5$ iterations while the learning rate decays by half at every $1 \times 10^5$ iterations. We implemented our network on Pytorch framework with NVIDIA RTX A4000 GPUs. We also implement our network by using the MindSpore Lite tool\footnote{\url{https://www.mindspore.cn/}}.

\subsection{Ablation Studies}\label{subsection:4-B}
Here, we observe the best PSNR values in $1 \times 10^5$ iterations for a quick comparison with isotropic Gaussian blur degradation.

\begin{table}
	\centering
	\caption{PSNR(dB) performance and the model complexity with different G and B. The results are evaluated on the Set5 dataset with \emph{Gaussian8} kernels for $\times4$ SR.}
	\begin{tabular}{|c|c|c|}
		\hline
		Variants & Number of parameters (M) & PSNR \\
		\hline 
		G5B5&8.0&31.50\\
		\hline
		G4B8&9.0&31.61\\
		\hline
		G5B10&11.7&31.69\\
		\hline
		G8B8&14.1&31.70\\
		\hline
		G10B10&\textbf{19.7}&31.75\\
		\hline
	\end{tabular}
	\label{tab:1}
\end{table}

\textbf{Investigation of the numbers of RGs and MCBs.} Firstly, we investigate the basic parameter of the proposed CDCN: the numbers of RGs (denoted as G for short), and the numbers of MCBs in each RG (denoted as B for short). Specifically, we develop 5 variants with different values of G and B to study how G and B will affect the SR performance, which are denoted as G5B5, G4B8, G5B10, G8B8, and G10B10. We observe the PSNR performance on Set5 dataset with \emph{Gaussian8} kernels for $\times4$ SR. As shown in Table \ref{tab:1}, we can see the model ``G10B10'' achieves the best PSNR but also involves the largest number of parameters which is about twice as ``G5B10''. ``G5B10'' shows very close PSNR to ``G8B8'' but fewer parameters (11.7M \emph{v.s.} 14.1M, about 20\% drop). To pursue a good balance of model complexity and SR performance, we choose the variant G5B10 as our baseline model in other experiments.

\begin{table}[t]
	\centering  
	\caption{Effect of the component decomposition (PSNR(dB)) measured on the Set5 \cite{set5} for $ 4\times $ SR .The kernel width is set to 1.2.}
	\begin{adjustbox}{width=\columnwidth,center}
		\begin{tabular}{|@{}c@{}|c|c|c|c|c|}
			\hline
			& Model 1 &Model 2 & Model 3 & Model 4 & Model 5 \\
			\hline
			Decomposition & & \Checkmark & \Checkmark & \Checkmark & \Checkmark \\
			Structure Constraint & & & \Checkmark & & \Checkmark \\
			Detail Constraint & & & & \Checkmark &  \Checkmark \\
			\hline
			PSNR & 31.59 & 31.65 & 31.74 & 31.67 & \textbf{31.79} \\
			\hline
		\end{tabular}
	\end{adjustbox}
	\label{tab:2}
\end{table}

\textbf{Study of Components Decomposition.} As mentioned in Section \ref{section:3}, we decompose the input LR images into structure components and detail components. The network is then trained with the degradation-driven learning strategy, which makes supervision using the corresponding labels. 
In this subsection, we investigate the effect of our components decomposition and the constraints of both components.

Here, we first introduce a network variant (Model 1) by removing the component decomposition module in Fig.~\ref{fig:2}. Specifically, the proposed dual-path network is simplified to a single-path network. Then, we develop another variant by removing both structure and detail constraints (Model 2). As shown in Table \ref{tab:2}, the network without components decomposition shows the worst PSNR score among all variants. Obviously, our CDCN benefits from both components decomposition and constraints from structure and detail components to produce better results.

To further investigate which component plays a more important role in the proposed network, we develop two variants by removing structure constraint (Model 3) or detail constraint (Model 4). From Table \ref{tab:2}, it can be found that only structure constraint or only detail constraint is not enough to recover satisfactory HR images. Models 3 and 4 suffer from the absence of component constraint, thus producing worse results compared to the full model. Such a phenomenon demonstrates the importance of component learning to a certain extent. It is worth noting that, they still produce better results than Model 1 and 2, which proves that the absence of component constraint will not destroy the whole network recovering ability. 

\begin{figure}[t]
	\centering
	\includegraphics[width=1.0\linewidth]{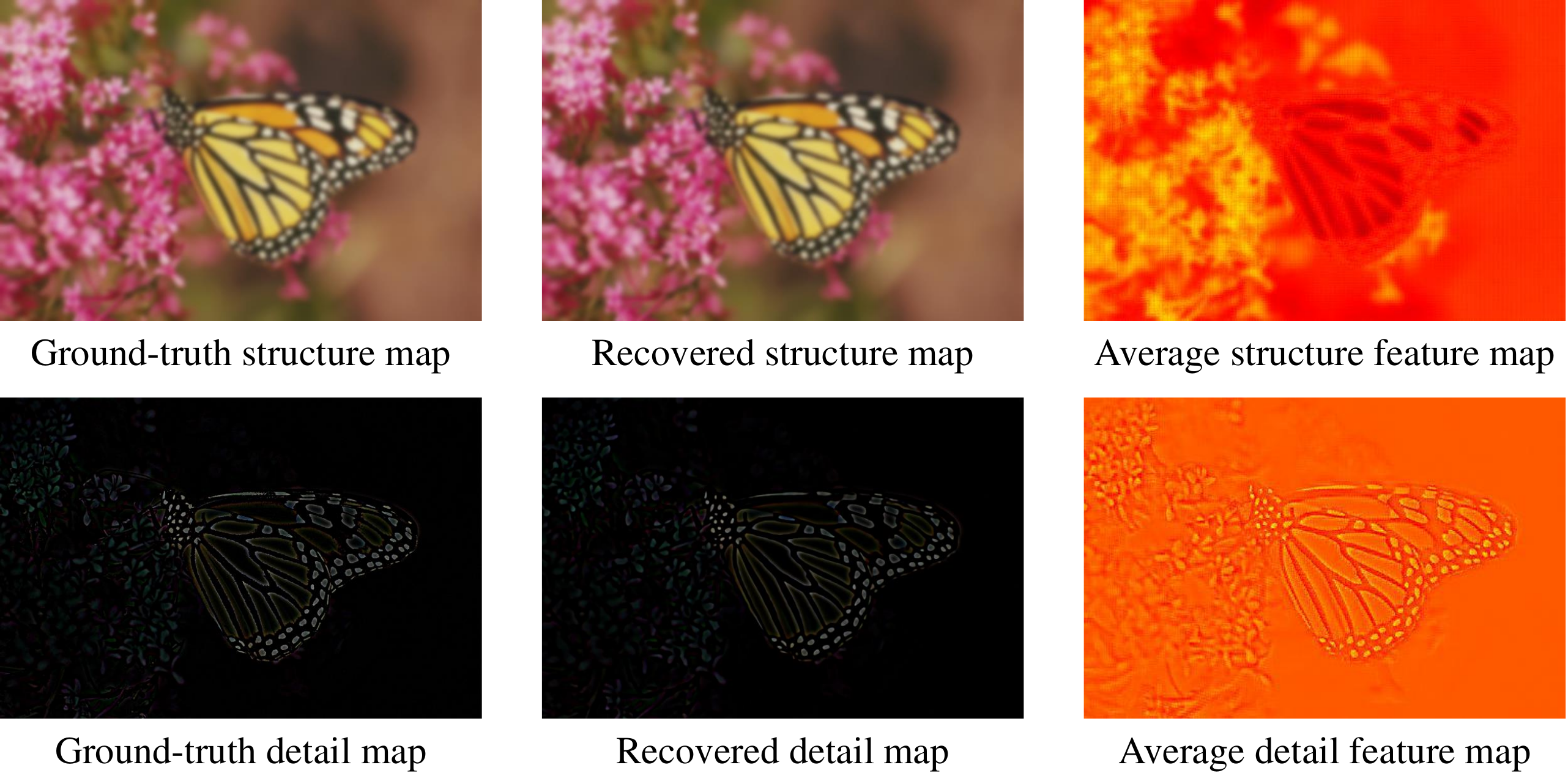}
	\caption{Visualized recovered detail and structure components of the proposed CDCN with isotropic kernel for $\times4$ SR. The kernel width is set to 3.6.}
	\label{fig:5}
\end{figure}

In Fig.~\ref{fig:5}, we visualize the recovered structure and detail maps and their corresponding feature maps to illustrate how well our structure and detail components have been learned by the proposed CDCN. Noted that these feature maps are produced by the last convolutional layers before the final reconstruction layers. We can get two intuitive clues from this visualization: (1) Our framework can recover promising structure and detail maps that are similar to their corresponding ground-truth versions (first column). These comparisons demonstrate that the proposed CDCN can effectively exploit the interaction between the image structure and detail components for better HR reconstruction. (2) The average structure feature map before reconstruction tends to contain more low-frequency structural information, while the average detail feature map has more discriminative high-frequency detail information (\emph{e.g.} edges and textures). Such a phenomenon shows a strong effect of structure and detail constraints, which further leads to a more accurate recovered structure or detail map. With the well-recovered components, our framework can achieve better SR reconstruction.

In the next subsection, we will investigate how the collaborative optimization of structure and detail components affects the final SR performance.

\begin{table}[h]
	\centering
	\caption{Effect of the variants (PSNR(dB)/SSIM) in our MCB measured on the Set14 $\times 4$ dataset.}
	\begin{tabular}{|c|ccc|}
		\hline
		\multirow{2}{*}{Variant} & \multicolumn{3}{c|}{Kernel Width}  \\
		& 1.2 & 2.4 & 3.6 \\
		\hline
		w/o all & 28.26/0.7670 & 28.10/0.7598 & 26.98/0.7147  \\ 
		w/o CO & 28.34/0.7700 & 28.16/0.7623 & 26.88/0.7112 \\
		w/ FC & 28.39/0.7726 & 28.15/0.7612 & 27.13/0.7175 \\
		w/ EA & 28.37/0.7731 & 28.15/0.7615 & 27.12/0.7198 \\
		w/ all & \textbf{28.50}/\textbf{0.7777} & \textbf{28.29}/\textbf{0.7670} & \textbf{27.24}/\textbf{0.7239} \\
		\hline
	\end{tabular}
	\label{tab:3}
\end{table}

\begin{figure}[h]
	\centering
	\includegraphics[width=1.0\linewidth]{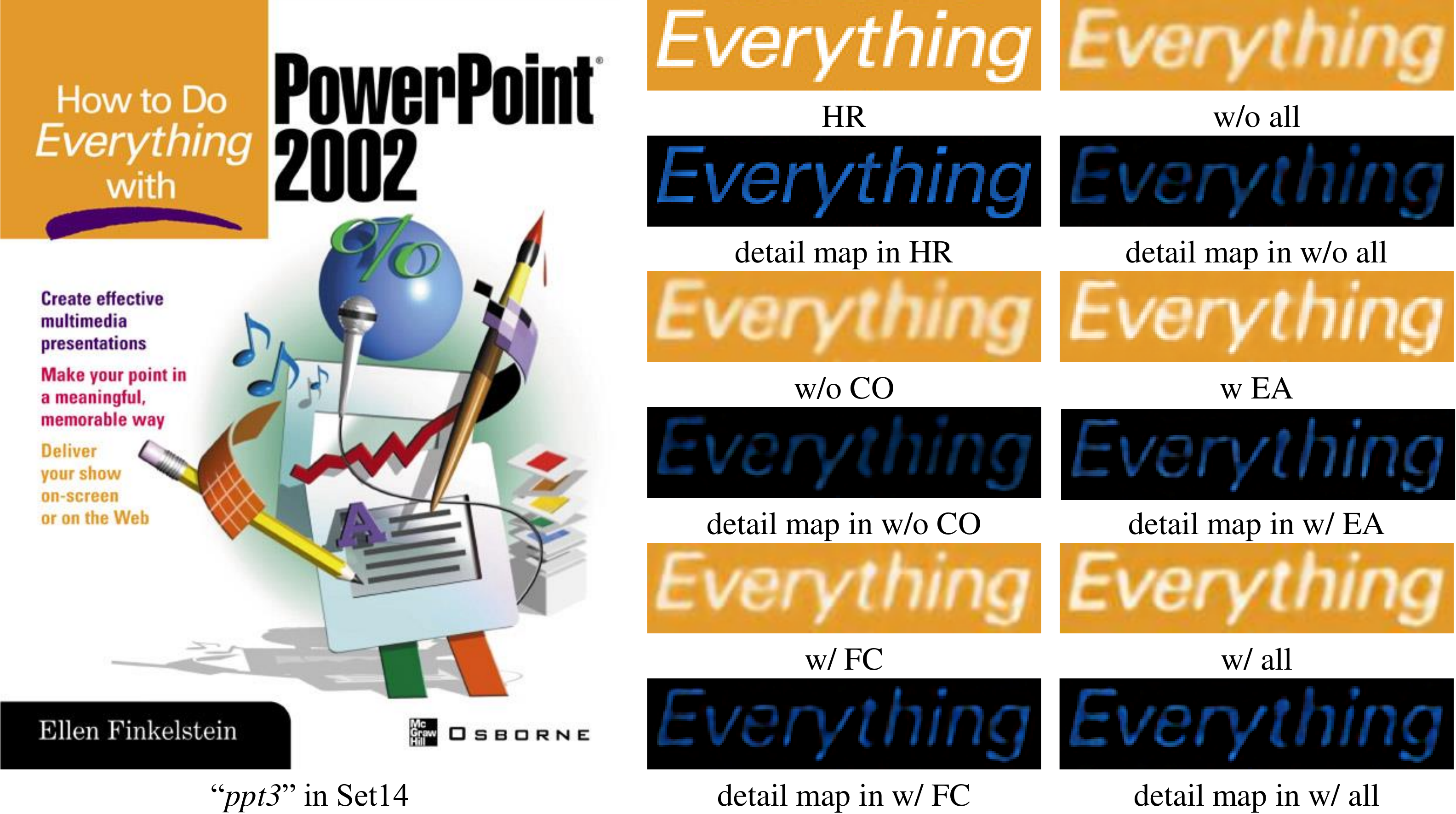}
	\caption{Visual comparisons with different variants with isotropic Gaussian kernels on Set14 \cite{set14} for $\times 4$ SR. The kernel width is set to 3.6.}
	\label{fig:6}
\end{figure}

\begin{figure}[h]
	\centering
	\includegraphics[width=1.0\linewidth]{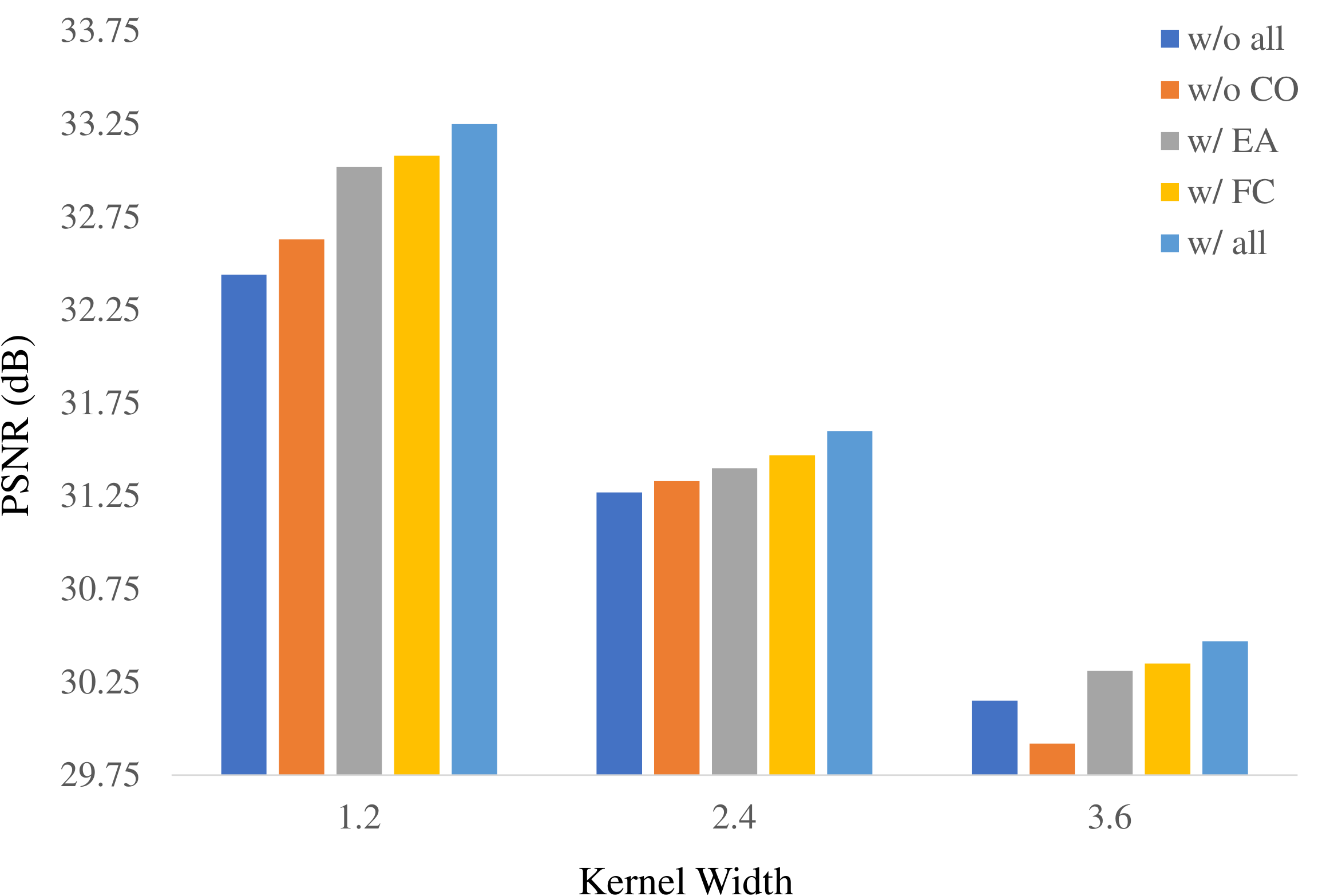}
	\caption{PSNR results of detail estimation with different kernel width on Set14 \cite{set14} for $\times 4$ SR.}
	\label{fig:7}
\end{figure}

\textbf{Influence of Collaborative Optimization} With the proposed MCB, we achieve the collaborative optimization of structure and detail components. Here, we investigate the effect of our proposed MCB by comparing our model with another 4 variants. First, we remove feature concatenation and channel attention operation, which simplifies our model to a simple dual-path residual block (w/o all). Then, we keep the channel attention operation and introduce a variant without collaborative optimization (w/o CO), which makes our residual group constructed by dual-path residual channel-attention blocks. In that situation, the whole residual group learns the two components separately and only fuses them in the final MSFM. To further investigate the components fusion and interaction ability of the proposed MCB, We remove the channel attention operation and directly use the feature concatenation operation with $1 \times 1$ convolution to achieve component fusion (w/ FC). hen, we replace feature concatenation operation with element-wise addition where the features from another component are fed into $1 \times 1$ convolution and then added to the original component (w/ EA). 

As we can see from Table \ref{tab:3}, the channel attention operation improves the performance to some extent, but it fails in certain kernel widths. The variants which utilize collaboration between the structure and detail components ``w/ FC'' and ``w/ EA'' achieve better performance than those without information interaction ``w/o all'' and ``w/o CO''. Compared with other variants, our model constructed with the proposed MCB can effectively exploit the relationship between detail and structure components for better SR performance. 

To further demonstrate the effectiveness of MCB, we also visualize the detail components from the variants (see Fig.~\ref{fig:6}) and calculate PSNR values (see Fig.~\ref{fig:7}) between the detail components and their corresponding ground-truths. It can be observed in Fig. \ref{fig:6} that the proposed MCB recovers brighter and cleaner detail components than other variants, which directly leads to better SR reconstruction in the texts \emph{``Everything"}. Fig.~\ref{fig:7} demonstrates the effectiveness of MCB for detail estimation from a quantitative perspective.

\begin{table}[h]
	\centering
	\caption{Effect of the variants (PSNR(dB)) in our Multi-Scale Fusion Module measured on the Set5 \cite{set5}, Set14 \cite{set14} for $ 4\times $ SR .The kernel width is set to 2.4.}
	\begin{tabular}{|c|cc|}
		\hline
		\multirow{2}{*}{Variant} & \multicolumn{2}{c|}{Dataset}  \\
		& Set5 & Set14 \\
		\hline
		w/o MSFM & 31.46 & 28.04  \\ 
		w/o multi-scale feature fusion & 31.62 & 28.08  \\
		w/o fused feature enhancement & 31.52 & 28.16  \\
		w/ MSFM & \textbf{31.65} & \textbf{28.29}  \\
		\hline
	\end{tabular}
	\label{tab:4}
\end{table}

\begin{table*}[h]
	\centering
	\caption{Quantitative comparisons (PSNR(dB)) with SOTA blind SR methods with \emph{Gaussian8} kernels on Set5 \cite{set5}, Set14 \cite{set14}, BSD100 \cite{b100}, Urban100 \cite{urban100} and Manga109 \cite{manga109}. The best two results are highlighted in \textcolor{red}{red} and \textcolor{blue}{blue} colors.}
	\resizebox{\linewidth}{!}{
		\begin{tabular}{cccccccccccc}
			\toprule
			\multirow{2}{*}{Method} & \multirow{2}{*}{Scale} & \multicolumn{2}{c}{Set5} & \multicolumn{2}{c}{Set14} & \multicolumn{2}{c}{BSD100} & \multicolumn{2}{c}{Urban100} & \multicolumn{2}{c}{Manga109}\\
			& & PSNR & SSIM & PSNR & SSIM & PSNR & SSIM & PSNR & SSIM & PSNR & SSIM \\
			\midrule
			Bicubic & \multirow{10}{*}{$\times2$} & 28.82 & 0.8577 & 26.02 & 0.7634 & 25.92 & 0.7310 & 23.14 & 0.7528 & 25.60 & 0.8498 \\
			RCAN \cite{RCAN}& & 31.37 & 0.8918 & 28.63 & 0.8144  & 28.16 & 0.7822 & 25.58 & 0.7863 & 28.48 & 0.8933 \\
			CARN \cite{CARN}& & 30.99 & 0.8779 & 28.10 & 0.7879 & 26.78 & 0.7286 & 25.27 & 0.7630 & 26.86 & 0.8606 \\
			Bicubic + ZSSR \cite{ZSSR} & & 31.08 & 0.8786 & 28.35 & 0.7933 & 27.92 & 0.7632 & 25.25 & 0.7618 & 28.05 & 0.8769 \\
			\cite{deblur} + CARN \cite{CARN} & & 24.20 & 0.7496 & 21.12 & 0.6170 & 22.69 & 0.6471 & 18.89 & 0.5895 & 21.54 & 0.7946 \\
			CARN \cite{CARN} + \cite{deblur} & & 31.27 & 0.8974 & 29.03 & 0.8267 & 28.72 & 0.8033 & 25.62 & 0.7981 & 29.58 & 0.9134 \\
			IKC \cite{IKC}& &  36.82 & \textcolor{blue}{0.9534} & 32.81 & \textcolor{blue}{0.9074} & 31.62 & 0.8884 & 30.21 & 0.9019 & 36.15 & 0.9660 \\
			DAN \cite{DAN}& & 37.34 & 0.9526 & 33.08 & 0.9041 &31.76 & 0.8858 & 30.60 & 0.9060 & 37.23 & 0.9710 \\
			DASR \cite{DASR}& & 37.01 & 0.9504 & 32.61 & 0.8959 & 31.59 & 0.8815 & 30.25 & 0.9016 & 36.20 & 0.9687 \\
			DSSR \cite{DSSR}& & \textcolor{blue}{37.46} & 0.9529 & \textcolor{blue}{33.17} & 0.9044 & \textcolor{blue}{31.95} & \textcolor{blue}{0.8901} & \textcolor{blue}{31.03} & \textcolor{blue}{0.9118} & \textcolor{blue}{37.40}& \textcolor{blue}{0.9720} \\
			CDCN (ours) & & \textcolor{red}{37.62} & \textcolor{red}{0.9550} & \textcolor{red}{33.45} & \textcolor{red}{0.9084} & \textcolor{red}{32.02} & \textcolor{red}{0.8905} & \textcolor{red}{31.44} & \textcolor{red}{0.9164} & \textcolor{red}{37.57} & \textcolor{red}{0.9729} \\
			\midrule
			Bicubic & \multirow{10}{*}{$\times3$} & 26.21 & 0.7766 & 24.01 & 0.6662 & 24.25 & 0.6356 & 21.39 & 0.6203 & 22.98 & 0.7576 \\
			RCAN \cite{RCAN}& & 28.28 & 0.8158 & 26.13 & 0.7133 & 26.06 & 0.6772 & 23.31 & 0.6750 & 25.16 & 0.8057 \\
			CARN \cite{CARN}& & 27.26 & 0.7855 & 25.06 & 0.6676 & 25.85 & 0.6566 & 22.67 & 0.6323 & 23.84 & 0.7620 \\
			Bicubic + ZSSR \cite{ZSSR} & & 28.25 & 0.7989 & 26.11 & 0.6942 & 26.06 & 0.6633 & 23.26 & 0.6534 & 25.19 & 0.7914 \\
			\cite{deblur} + CARN \cite{CARN} & & 19.05 & 0.5226 & 17.61 & 0.4558 & 20.52 & 0.5331 & 16.72 & 0.4578 & 18.38 & 0.6118 \\
			CARN \cite{CARN} + \cite{deblur} & & 30.31 & 0.8562 & 27.57 & 0.7532 & 27.14 & 0.7152 & 24.45 & 0.7241 & 27.67 & 0.8592 \\
			IKC \cite{IKC}& & 33.06 & 0.9146 & 29.38 & 0.8233 & 28.53 & 0.7899 & 27.43 & 0.8302 & 32.43 & 0.9316 \\
			DAN \cite{DAN}& & 34.04 & \textcolor{blue}{0.9199} & \textcolor{blue}{30.09} & \textcolor{blue}{0.8287} & 28.94 & 0.7919 & \textcolor{blue}{27.65} & \textcolor{blue}{0.8352} & \textcolor{blue}{33.16} & 0.9382 \\
			DASR \cite{DASR}& & 33.48 & 0.9125 & 29.64 & 0.8154 & 28.63 & 0.7829 & 26.88 & 0.8149 & 32.03 & 0.9286 \\
			DSSR \cite{DSSR}& & \textcolor{blue}{34.05} & 0.9197 & \textcolor{blue}{30.09} & 0.8270 & \textcolor{blue}{28.98} & \textcolor{blue}{0.7923} & 27.64 & 0.8349 & 33.07 & \textcolor{blue}{0.9384} \\
			CDCN (ours) & & \textcolor{red}{34.12} & \textcolor{red}{0.9209} & \textcolor{red}{30.19} & \textcolor{red}{0.8313} & \textcolor{red}{29.04} & \textcolor{red}{0.7952} & \textcolor{red}{27.91} & \textcolor{red}{0.8410} & \textcolor{red}{33.30} & \textcolor{red}{0.9400} \\
			\midrule
			Bicubic & \multirow{10}{*}{$\times4$} & 24.57 & 0.7108 & 22.79 & 0.6032 & 23.29 & 0.5786 & 20.35 & 0.5532 & 21.50 & 0.6933\\
			RCAN \cite{RCAN}& & 26.59 & 0.7591 & 24.84 & 0.6529 & 25.01 & 0.6197 & 22.19 & 0.6094 & 23.52 & 0.7449 \\
			CARN \cite{CARN}& & 26.57 & 0.7420 & 24.62 & 0.6226 & 24.79 & 0.5963 & 22.17 & 0.5865 & 21.85 & 0.6834 \\
			Bicubic + ZSSR \cite{ZSSR} & & 26.45 & 0.7279 & 24.78 & 0.6268 & 24.97 & 0.5989 & 22.11 & 0.5805 & 23.53 & 0.7240 \\
			\cite{deblur} + CARN \cite{CARN} & & 18.10 & 0.4843 & 16.59 & 0.3994 & 18.46 & 0.4481 & 15.47 & 0.3872 & 16.78 & 0.5371 \\
			CARN \cite{CARN} + \cite{deblur} & & 28.69 & 0.8092 & 26.40 & 0.6926 & 26.10 & 0.6528 & 23.46 & 0.6597 & 25.84 & 0.8035 \\
			IKC \cite{IKC}& & 31.67 & 0.8829 & 28.31 & 0.7643 & 27.37 & 0.7192 & 25.33 & 0.7504 & 28.91 & 0.8782 \\
			DAN \cite{DAN}& & 31.89 & 0.8864 & 28.35 & 0.7666 & \textcolor{blue}{27.51} & \textcolor{blue}{0.7248} & \textcolor{blue}{25.86} & \textcolor{blue}{0.7721} & \textcolor{blue}{30.50} & \textcolor{red}{0.9037}  \\
			DASR \cite{DASR}& & 31.57 & 0.8808 & 28.14 & 0.7603 & 27.35 & 0.7196 & 25.34 & 0.7543 & 29.79 & 0.8928 \\
			AdaTarget \cite{DSSR}& & 31.58 & 0.8814 & 28.14 & 0.7626 & 27.43 & 0.7216 & 25.72 & 0.7683 & 29.97 & 0.8955 \\
			SRDRL \cite{SRDRL}& & 29.06 & 0.8312 & 26.55 & 0.7118 & 26.28 & 0.6742 & 23.59 & 0.6769 & 25.86 & 0.8126 \\
			DSSR \cite{DSSR}& & \textcolor{blue}{31.97} & \textcolor{blue}{0.8870} & \textcolor{blue}{28.43} & \textcolor{blue}{0.7679} & 27.49 & 0.7229 & 25.73 & 0.7668 & 30.44 & 0.9023 \\
			CDCN (ours) & & \textcolor{red}{32.04} & \textcolor{red}{0.8883} & \textcolor{red}{28.47} & \textcolor{red}{0.7695} & \textcolor{red}{27.52} & \textcolor{red}{0.7257} & \textcolor{red}{25.92} & \textcolor{red}{0.7731} & \textcolor{red}{30.55} & \textcolor{blue}{0.9036}\\
			\bottomrule
		\end{tabular}   
	}
	\label{tab:5}
\end{table*}

\begin{figure*}[h]
	\centering
	\includegraphics[width=1.0\linewidth]{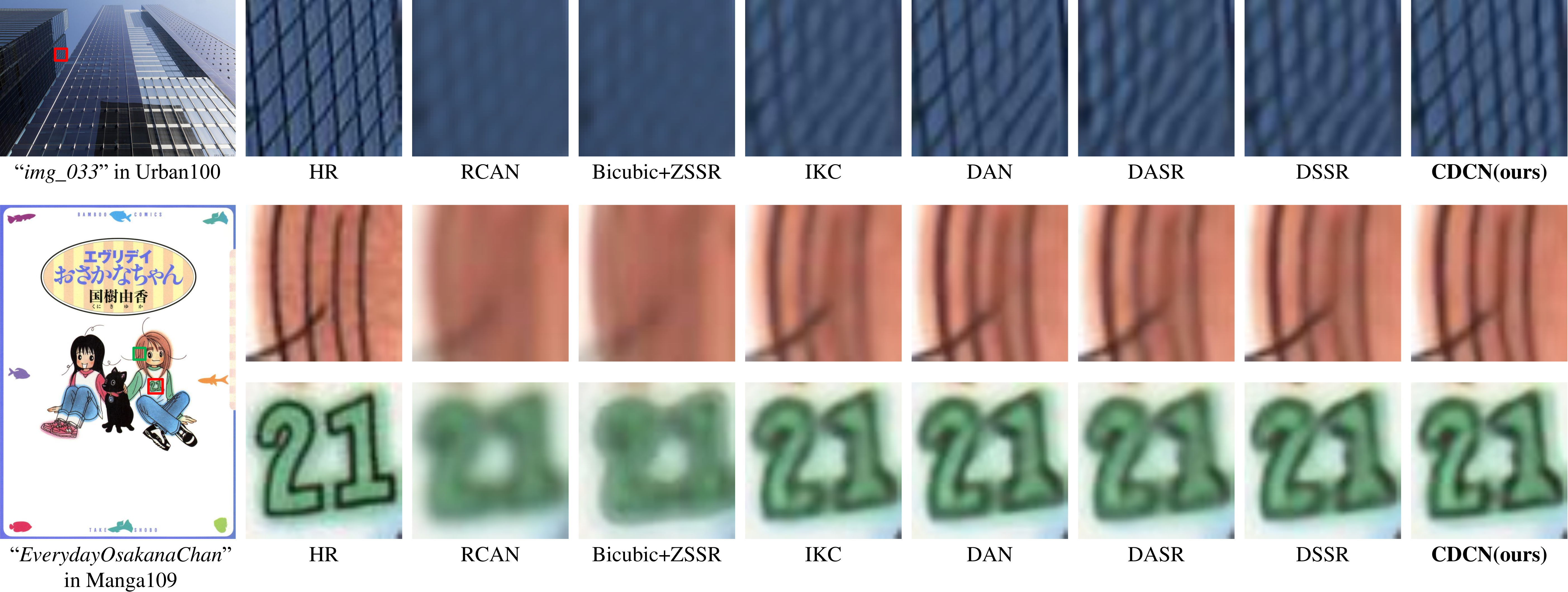}
	\caption{Qualitative comparisons with SOTA SR methods with isotropic Gaussian kernels on Urban100 \cite{urban100} and Manga109 \cite{manga109} for $\times 4$ SR. The kernel width is set to 2.4.}
	\label{fig:8}
\end{figure*}

\begin{table}[h]
	\centering
	\caption{Quantitative comparisons (PSNR(dB)) with SOTA non-blind SR methods with isotropic Gaussian kernels on Set5 \cite{set5}, Set14 \cite{set14} and BSD100 \cite{b100}. The kernel width is set to 1.3. The best two results are highlighted in \textcolor{red}{red} and \textcolor{blue}{blue} colors.}
	\begin{tabular}{|c|c|c|c|c|}
		\hline
		Methods & Scale & Set5 & Set14 & BSD100 \\
		\hline
		\hline
		SRMDNF \cite{SRMD}& \multirow{5}{*}{$\times2$} & 37.44 & 33.20 & 31.98 \\
		SFTMD w/o SFT \cite{IKC} & & 30.88 & 27.16 & 26.84 \\
		SFTMD \cite{IKC} & & \textcolor{blue}{37.46} & \textcolor{red}{33.39} & \textcolor{red}{32.06} \\
		MZSR \cite{MZSR} & & 36.07 & 32.15 & 30.94 \\
		DAN \cite{DAN} & & 37.32 & 33.07 & 31.76 \\
		UDVD \cite{UDVD} & & 37.36 & \textcolor{red}{33.39} & \textcolor{blue}{32.00} \\
		CDCN (ours) & & \textcolor{red}{37.58} & \textcolor{blue}{33.21} & 31.94 \\
		\hline
		SRMDNF \cite{SRMD}& \multirow{5}{*}{$\times3$} & 34.17 & 30.08 & 29.03 \\
		SFTMD w/o SFT \cite{IKC} & & 30.33 & 25.84 & 25.92 \\
		SFTMD \cite{IKC} & & \textcolor{blue}{34.53} & \textcolor{red}{30.55} & 29.15 \\
		DAN \cite{DAN} & & 34.46 & 30.33 & 29.13 \\
		UDVD \cite{UDVD} & & 34.52 & \textcolor{blue}{30.50} & \textcolor{red}{29.23} \\
		CDCN (ours) & & \textcolor{red}{34.56} & 30.45 & \textcolor{blue}{29.21} \\
		\hline
		SRMDNF \cite{SRMD}& \multirow{5}{*}{$\times4$} & 32.00 & 28.42 & 27.53 \\
		SFTMD w/o SFT \cite{IKC} & & 29.11 & 25.93 & 26.20 \\
		SFTMD \cite{IKC} & & \textcolor{red}{32.41} & \textcolor{blue}{28.82} & 27.64 \\
		DAN \cite{DAN} & & 32.19 & 28.65 & 27.66 \\
		UDVD \cite{UDVD} & & \textcolor{blue}{32.37} & \textcolor{red}{28.85} & \textcolor{red}{27.75} \\
		CDCN (ours) & & 32.32 & 28.64 & \textcolor{blue}{27.68} \\
		\hline
	\end{tabular}
	\label{tab:6}
\end{table}

\textbf{Effect of Multi-scale Fusion Module.} In the previous subsection, we prove the necessity of component decomposition and collaborative learning. Then we investigate the effect of the proposed multi-scale fusion module. We first introduce a variant by removing the MSFM in Fig.~\ref{fig:2}, where we simply use element-wise addition to put structure and detail features together and get the fused features for SR reconstruction. Then we develop another two variants by removing the multi-scale feature fusion stage or fused feature enhancement stage in Fig.~\ref{fig:4}. From Table \ref{tab:4}, it can be observed that the PSNR values decrease if two components are not well integrated (\emph{e.g.} 28.29 vs 28.04 for dataset Set14). Besides, both multi-scale feature fusion and fused feature enhancement play an important role in higher PSNR scores.

\subsection{Comparison with State-of-the-arts}

\textbf{Isotropic Gaussian Kernels.} For the isotropic Gaussian blur kernels, we first evaluate our method on synthesized test images by \emph{Gaussian8} kernel setting. We compare our method with state-of-the-art blind SR approaches including ZSSR \cite{ZSSR} (with bicubic kernel), IKC \cite{IKC}, DAN \cite{DAN}, DASR \cite{DAN}, AdaTarget \cite{adatarget}, SRDRL \cite{SRDRL}, DSSR \cite{DSSR}. We also conduct a comparison with bicubic-assumed methods RCAN \cite{RCAN}, CARN \cite{CARN}. Following \cite{DAN}, the deblurring method \cite{deblur} is performed before or after CARN to investigate the cooperation of deblurring methods and bicubic-assumed SR methods. For most methods, we use their public code and pre-trained models. Since the models of IKC for some scale factors are unavailable, we retrain IKC using its official implementation to evaluate the SR performance.

The quantitative comparison is shown in Table \ref{tab:5}. Although RCAN \cite{RCAN} and CARN \cite{CARN} achieve promising results on bicubic degradation, they suffer severe performance drop when the degradation deviates from the bicubic one since they are trained under the bicubic setting. When refined by the deblurring method (``CARN+\cite{deblur}"), the performance of CARN significantly improves and exceeds the large capacity network RCAN. However, when we perform the deblurring method before SR reconstruction (``\cite{deblur}+CARN"), the result is even worse than bicubic interpolation owing to the larger domain gap between deblurred images and degraded LR images.  Benefiting from correcting inaccurate blur kernels iteratively, IKC \cite{IKC} improves their results with higher PSNR values. By unifying degradation estimation and SR reconstruction into an end-to-end framework, DAN \cite{DAN} and DASR \cite{DASR} perform better than IKC, but they are still inferior to our method. Compared with DSSR \cite{DSSR}, which utilizes the same detail-structure optimization as us, our method outperforms the proposed degradation-driven learning strategy. 

In Table \ref{tab:6}, we even make a comparison to several non-blind SR methods, including SRMDNF \cite{SRMD}, SFTMD \cite{IKC}, MZSR \cite{MZSR}, UDVD \cite{UDVD}, to further show our performance on other specific kernel width. The result of SOTA blind SR approach DAN \cite{DAN} is also reported. As one can see, when the ground-truth blur kernels are available, non-blind SR methods perform better than blind SR methods, such as UDVD \emph{v.s.} DAN. Compared to these methods, even without blur kernel prior, the proposed CDCN not only produces better SR results than the blind SR method DAN, but also achieves comparable performance with existing non-blind SR methods. The comparison in Table \ref{tab:6} demonstrates the robustness of our method. 

The qualitative results are shown in Fig.~\ref{fig:8}. It can be observed that RCAN and ZSSR can hardly recover any details in the images, thus their results are still blurred. The recent SOTA blind SR methods like IKC, DAN, DASR, and DSSR produce cleaner SR images in general but fail in certain areas in the images (\emph{e.g.}, the lower right part of the building in \emph{``img\_033"} from Urban100 dataset in Fig.~\ref{fig:8}). In contrast to them, the proposed CDCN generates visually pleasant results with better color fidelity and higher perceptual quality.

\textbf{Anisotropic Gaussian Kernels.} For anisotropic Gaussian blur kernels, following \cite{kernelgan,DAN,DSSR}, we compare methods of four different classes:
\begin{itemize}
  \item [1)] 
  SR methods trained on bicubic downsampled images such as EDSR \cite{EDSR} and RCAN \cite{RCAN}.       
  \item [2)]
  Blind SR methods which win NTIRE blind SR challenge including WDSR \cite{wdsr} and RealSR \cite{realsr}.
  \item [3)]
  Kernel estimation methods such as KernelGAN \cite{kernelgan} combined with non-blind SR methods such as SRMD \cite{SRMD} and USRNet \cite{USRNet}.
  \item [4)]
  Blind SR methods integrating kernel estimation and SR reconstruction into an end-to-end framework such as IKC \cite{IKC}, DAN \cite{DAN} and KOALAnet \cite{KOALAnet}.
  \item [5)]
  Blind SR methods without kernel estimation such as DASR \cite{DASR} and DSSR \cite{DSSR}.
\end{itemize}

\begin{figure*}[h]
	\centering
	\includegraphics[width=1.0\linewidth]{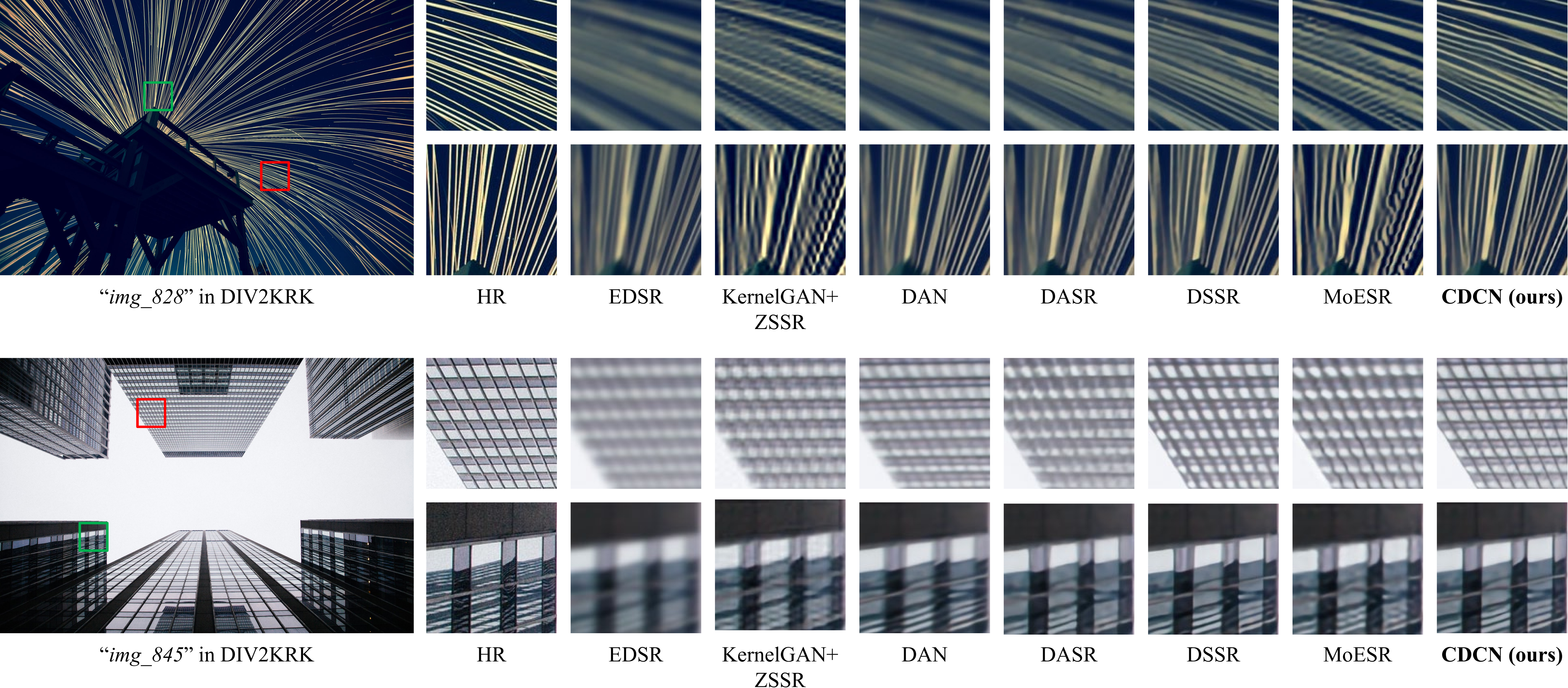}
	\caption{Qualitative comparisons with SOTA SR methods with anisotropic Gaussian kernels on DIV2KRK \cite{kernelgan} for $\times 4$ SR.}
	\label{fig:9}
\end{figure*}

\begin{table*}[h]
	\centering
	\caption{Quantitative comparisons (PSNR(dB)/SSIM) with SOTA SR methods with anisotropic Gaussian kernels on DIV2KRK \cite{kernelgan}. The best two results are highlighted in \textcolor{red}{red} and \textcolor{blue}{blue} colors.}
	\begin{tabular}{|c|c|c|c|}
		\hline
		\multirow{2}{*}{Type} & \multirow{2}{*}{Method} & \multicolumn{2}{c|}{Scale} \\
		\cline{3-4}
		& & $\times2$ & $\times4$ \\
		\hline
		\multirow{4}{*}{\textbf{Class 1}: SR methods trained on bicubic downsampled images} & Bicubic & 28.73 / 0.8040 & 25.33 / 0.6795 \\
		& Bicubic + ZSSR \cite{ZSSR} & 29.10 / 0.8215 & 25.61 / 0.6911 \\
		& EDSR \cite{EDSR} & 29.17 / 0.8216 & 25.64 / 0.6928 \\
		& RCAN \cite{RCAN} & 29.20 / 0.8223 & 25.66 / 0.6936 \\
		\hline
		\multirow{5}{*}{\textbf{Class 2}: winners of NTIRE \cite{ntire2018,ntire2020} blind SR challenge} & WDSR \cite{wdsr} - 1st in NTIRE' 18 \cite{ntire2018} track2 & - & 21.55 / 0.6841 \\
		& WDSR \cite{wdsr} - 1st in NTIRE' 18 \cite{ntire2018} track3 & - & 21.54 / 0.7016 \\
		& WDSR \cite{wdsr} - 2nd in NTIRE' 18 \cite{ntire2018} track4 & - & 25.64 / 0.7144 \\
		& RealSR \cite{realsr} - 1st in NTIRE' 20 \cite{ntire2020} track1 & - & 25.43 / 0.6907 \\
		\hline
		\multirow{5}{*}{\textbf{Class 3}: kernel estimation with non-blind methods} & Michaeli \emph{et al.} \cite{nonparametric} + SRMD \cite{SRMD} & 25.51 / 0.8083 & 23.34 / 0.6530 \\
		& Michaeli \emph{et al.} \cite{nonparametric} + ZSSR \cite{ZSSR} & 29.37 / 0.8370 & 26.09 / 0.7138 \\
		& KernelGAN \cite{kernelgan} + SRMD \cite{SRMD} & 29.57 / 0.8564 & 25.71 / 0.7265 \\
		& KernelGAN \cite{kernelgan} + ZSSR \cite{ZSSR} & 30.36 / 0.8869 & 26.81 / 0.7316 \\
		& KernelGAN \cite{kernelgan} + USRNet \cite{USRNet} & 23.81 / 0.6813 & 23.34 / 0.5370 \\
		\hline
		\multirow{5}{*}{\textbf{Class 4}: end-to-end blind SR methods based on kernel estimation} & BlindSR \cite{blindSR} & 29.44 / 0.8464 & - / - \\
		& IKC \cite{IKC} & 31.44 / 0.8817 & 27.70 / 0.7668 \\
		& KOALAnet \cite{KOALAnet} & 31.89 / 0.8852 & 27.77 / 0.7637 \\
		& DAN \cite{DAN} & 32.56 / 0.8997 & 27.55 / 0.7582 \\
		& MoESR \cite{moesr} & \textcolor{blue}{32.69} / \textcolor{red}{0.9054} & 28.48 / 0.7805 \\
		\hline 
		\multirow{3}{*}{\textbf{Class 5}: end-to-end blind SR methods without kernel estimation} & DASR \cite{DASR} & 31.07 / 0.8694 & 28.09 / 0.7687 \\
		& AdaTarget \cite{adatarget} & - / - & 28.42 / 0.7854 \\
		& DSSR \cite{DSSR} & 32.46 / 0.9018 & \textcolor{blue}{28.78} / \textcolor{blue}{0.7905} \\
		\hline
		Ours & CDCN  & \textcolor{red}{32.73} / \textcolor{blue}{0.9051} & \textcolor{red}{28.92} / \textcolor{red}{0.7929} \\
		\hline
	\end{tabular}
	\label{tab:7}
\end{table*}

As shown in Table \ref{tab:7}, the proposed CDCN maintains preferable performance across all scale factors compared with other methods. Specifically, since the methods in Class 1 are trained on bicubic downsampled images, they show the worst results on anisotropic Gaussian kernels as well as isotropic Gaussian kernels. The winners (Class 2) of NTIRE blind SR challenge achieve fine results in the challenge, but still fail to recover promising images in the face of anisotropic Gaussian kernels. When blur kernels are unavailable, due to the distance between estimated kernels and ground-truth kernels, simply combining kernel estimation methods and non-blind SR methods (Class 3) can not always produce satisfactory results. The end-to-end blind SR methods (Class 4) achieve superior results than other classes, but their performance is still limited by the accuracy of kernel estimation. Particularly, MoESR \cite{moesr} uses a mixture of experts for specific degradation kernels, which achieves remarkable performance for scale factor 2 but worse $4\times$ results. Some other methods conduct blind SR without kernel estimation (Class 5), which obtains close PSNR and SSIM to Class 4 but is still lower than the proposed CDCN. By comparison, our method solves the blind SR problem using detail and structure collaborative optimization with a degradation-driven learning strategy, which can achieve the almost best performance on all scale factors. 

Fig.~\ref{fig:9} shows the visual results in DIV2KRK \cite{kernelgan}. We can observe that the bicubic-based methods EDSR still fails to recover sharp lines. The SR images generated by KernelGAN and MoESR suffer from severe unpleasant artifacts. The SR images produced by DAN are cleaner, but visible artifacts and blurs yet exist in their result. Meanwhile, the proposed CDCN reduces the artifacts and recovers photo-realistic details like the window in \emph{``img\_845"} from DIV2KRK dataset in Fig.~\ref{fig:9}.

\begin{figure}[t]
	\centering
	\includegraphics[width=1.0\linewidth]{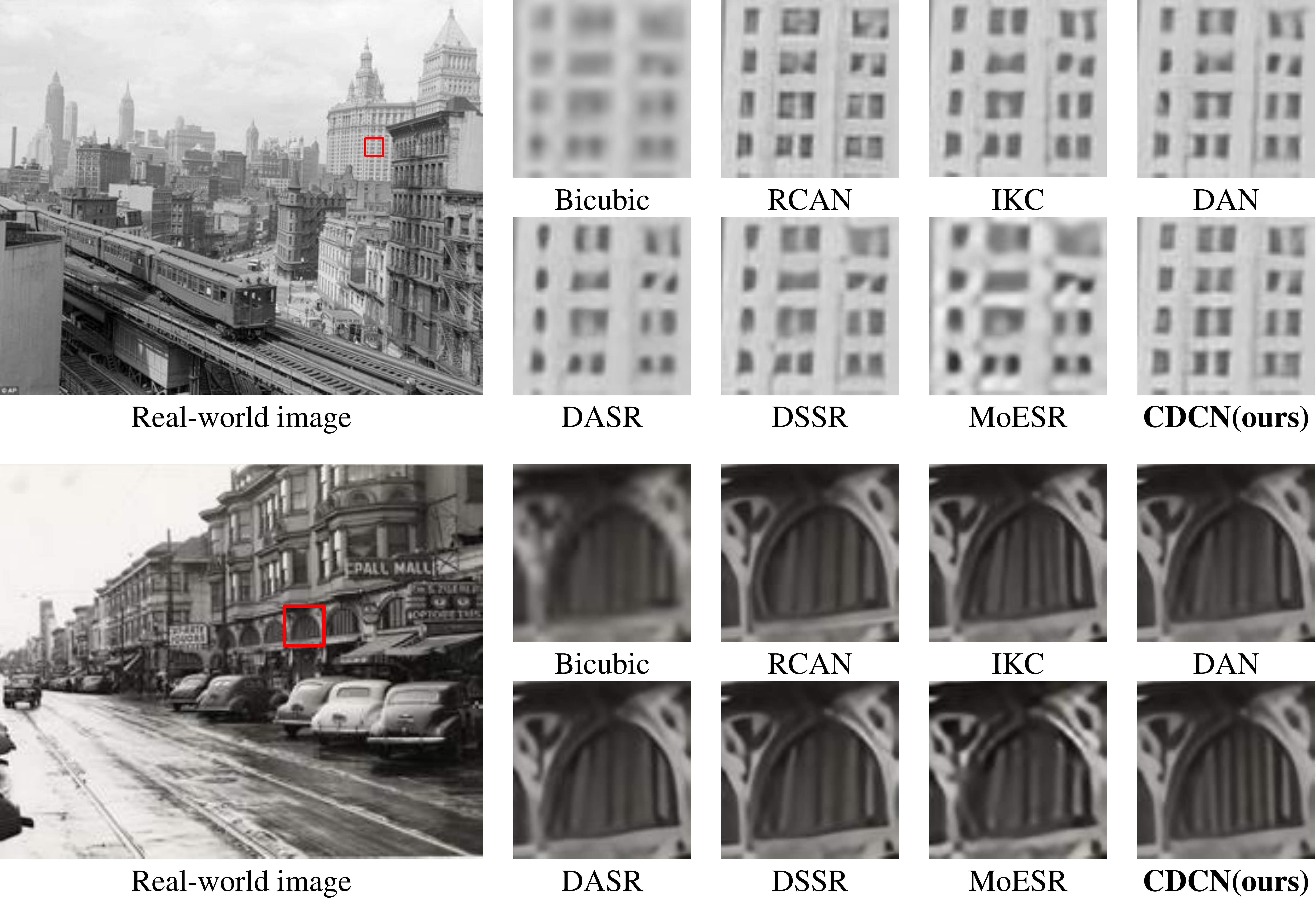}
	\caption{Qualitative comparisons with SOTA SR methods on real-world images for $\times 4$ SR.}
	\label{fig:10}
\end{figure}

\textbf{Real-World degradations.} In addition to experiments on the synthetic data set based on isotropic and anisotropic Gaussian kernels, we further compare with other methods on real-world images to demonstrate the effectiveness of the proposed CDCN. The visual results are shown in Fig. \ref{fig:10}. As one can see, the results of other compared methods are better than bicubic interpolation. No matter the existing non-blind SR or blind SR methods, they all super-resolve the images with obvious artifacts and blurry edges. Compared with these methods, our method produces SR images with sharper edges, clearer contents, and fewer artifacts, which demonstrates the generalization ability of the proposed CDCN.

\begin{table*}[h] 
	\renewcommand\arraystretch{}
	\caption{The efficiency comparison with several SOTA blind SR methods for $\times2$ SR on set5 dataset with \emph{Gaussian8} kernels.}
	\label{tab:8}
	\centering
	\begin{tabular}{|c| c| c| c| c| c| c | c|}
		\hline
		Methods & Bicubic+ZSSR\cite{ZSSR} & \cite{kernelgan}+ZSSR \cite{ZSSR} & IKC \cite{IKC} & DAN \cite{DAN} & DASR \cite{DASR} & DSSR\cite{DSSR} & CDCN (ours) \\
		\hline
		PSNR(dB) & 31.08	& 31.22&	36.82&	37.34&	37.01&	37.46 &\textbf{37.62}\\
		\hline
		Parameters & 2.2M &0.18M+2.2M & 5.2M & 4.3M	& 5.8M	&8.9M	&11.7M\\	
		\hline
		Time(sec.) & 28.51	&229.03&	3.01&	0.55&\textbf{0.13}&0.56&	0.19\\
		\hline
	\end{tabular}
\end{table*}
\textbf{Efficiency.} To demonstrate the efficiency of the proposed CDCN, we conduct a comparison with other SOTA blind SR methods in terms of parameters, average inference time, and PSNR performance on Set5 dataset with \emph{Gaussian8} kernels for $\times2$ SR. For fair comparison, all the methods are tested using their public source codes and pre-train models on the same platform TITAN RTX GPU. As shown in Table \ref{tab:8}, although having the fewest parameters, the zero-shot methods ``Bicubic + ZSSR'' and ``\cite{kernelgan} +ZSSR'' suffer from thousands of iterations during testing and show the worst inference time. Benefiting from iterative kernel estimation, IKC \cite{IKC} and DAN \cite{DAN} achieve better performance than ZSSR. The performance of DSSR \cite{DSSR} is further improved by utilizing the detail-structure optimization, but still lower than ours. Although the number of parameters of these iterative methods is fewer than the proposed CDCN, the inference time of these methods is much higher than that of ours due to their iteration strategy. Compared to the fastest method DASR \cite{DASR}, our method keeps a good balance between the inference time and the reconstruction performance. Specifically, our method exceeds DASR by 0.61 dB with slightly a higher execution time.

\section{Conclusion}
In this paper, we start from a degradation-based formulation model and analyze the degradation process from the perspective of image components. On the basis of degradation modelling, we propose a components decomposition and co-optimization network (CDCN), which performs simultaneously structure and detail components decomposition and restoration. In particular, CDCN uses the proposed mutual collaboration block (MCB) to extract features from both components and exploit the mutual relationship between them. Furthermore, a degradation-driven learning strategy is proposed to explicitly supervise the detail and structure restoration and exploit the interaction between the degradation kernel and SR reconstruction. Finally, a multi-scale fusion module is designed for enhancing the feature representations of both components and further producing a comprehensive feature for SR reconstruction. Extensive experiments on both synthetic datasets and real-world images demonstrate the effectiveness of the proposed CDCN. In the future, we will continue to mine the internal information of images and combine it with the degradation process. By utilizing internal image information analysis and external dataset training, we believe there will be further improvement in the blind SR task.  

\bibliographystyle{IEEEtran}
\bibliography{CDCN}

\end{document}